\documentclass[11pt, breaklinks,backref]{article} 

\usepackage[preprint]{ACL/acl}
\usepackage{cuted} 

\usepackage{times}
\usepackage{latexsym}

\usepackage[T1]{fontenc}

\usepackage[utf8]{inputenc}

\usepackage{microtype}

\usepackage{inconsolata}

\usepackage[utf8]{inputenc} 
\usepackage[T1]{fontenc}    
\usepackage{tgtermes}
\usepackage{url}            
\usepackage{booktabs}       
\usepackage{amsfonts}       
\usepackage{nicefrac}       
\usepackage{microtype}      
\usepackage{amsthm}
\usepackage{graphicx}
\usepackage{amsmath}
\usepackage{amssymb}
\usepackage{xr}
\usepackage{hyphenat}
\usepackage{subcaption}
\usepackage{float}
\usepackage{array}
\usepackage[table,xcdraw]{xcolor}
\usepackage{siunitx}
\usepackage{multirow} 
\usepackage{subcaption}
\usepackage{helvet}

 \renewcommand*{\backrefalt}[4]{%
    \ifcase #1%
     \or (Back to Section~#2)%
     \else (Back to Sections: ~#2)%
    \fi%
    }

\usepackage[shortlabels]{enumitem}
\setlist[itemize]{leftmargin=*}
\setlist[enumerate]{leftmargin=*}
\usepackage[normalem]{ulem}

\title{Vision-Language Models Mistake Head Orientation \\for Gaze Direction: Nonverbal Conversation Cues}

\author{
  {\bf Zory Zhang$^{\dagger 1}$}, {\bf Pinyuan Feng$^{\dagger 2}$}, {\bf Bingyang Wang$^3$}, {\bf Tianwei Zhao$^4$}, {\bf Suyang Yu$^5$}, \\
  {\bf Qingying Gao$^4$}, {\bf Hokin Deng$^{* 6}$}, {\bf Ziqiao Ma$^{* 7}$}, {\bf Yijiang Li$^{* 8}$}, {\bf Dezhi Luo$^{* 7}$} \\\\
  $^1$Brown University, $^2$Columbia University, $^3$Emory University, $^4$Johns Hopkins University, \\
  $^5$University of Washington, $^6$Carnegie Mellon University, $^7$University of Michigan, $^8$UC San Diego \\
    \raisebox{-0.4\height}{\includegraphics[height=6ex]{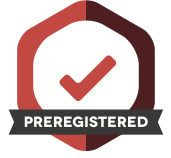}}
    \hspace{0.3em}
    \raisebox{-0.0\height}{
      \href{https://zoryzhang.github.io/gaze/}{Project Page} zoryzhang.github.io/gaze | \href{https://huggingface.co/datasets/Zory/gaze-referent-stimuli}{Dataset} | 
      \href{https://github.com/zoryzhang/referential-gaze}{Analysis Code}
    }
}

\begin{document}

\maketitle

\def\thefootnote{$\dagger$}\footnotetext{Equal Contribution. Correspondence: {zory@brown.edu}}
\def\thefootnote{*}\footnotetext{Equal Advising.}
\def\thefootnote{\arabic{footnote}}

\begin{abstract}
Where someone looks is a nonverbal communication cue that children and adults readily use.
How well can Vision-Language Models (VLMs) infer gaze targets? To construct evaluation stimuli, we captured 1,360 real-world photos of scenes in which a person gazes at one of several objects on a table. Importantly, we also controlled the gazer's head orientation: sometimes it was directed toward the gaze target, sometimes toward a distractor object, and sometimes left unconstrained. 
We found a substantial performance gap between VLMs and humans, ruled out alternative explanations such as resolution and object-naming skills, and identified the main reason for the gap as VLMs inferring gaze direction using head orientation rather than eye appearance.
Such a bias is likely due to data rather than architecture, as suggested by a proof-of-concept experiment finetuning a transformer-based vision model.
Future work should investigate whether these findings hold broadly across various deep learning methods trained on existing data, and whether better data mitigates this problem for all architectures.
Pinpointing the reason sets the stage for technologies that can interpret gaze targets to have more efficient interactions with humans.
\end{abstract}
\section{Introduction}
Gaze is a nonverbal communication channel that grounds verbal communication in the visual world, making it essential for both humans and technologies designed to engage with humans.
It contains information about conversational partners' unconscious knowledge and mental states, including their visual focus \citep{attention_sys}, linguistic knowledge \citep{langacquisition_gaze}, intention to speak during conversation turn-taking \citep{turntaking_2023}, and intended motor actions \citep{eye_motor}.
For conversational robots specifically, when they resolve what a conversational partner is referring to, gaze compensates for incomplete knowledge of the conversation topic \citep{prasov2008s} and differing perceptions of the shared environment \citep{gaze_mismatch_env}, thereby leading to more successful cooperation \citep{referent_robot}.

Nonetheless, integrating gaze inference into interactions is computationally challenging and involves a trade-off between specialization and generality.
Gaze-related skills can be divided into (1) the stepping stone of recognizing where someone is looking and (2) downstream skills that build on this foundation, like resolving conversational reference ambiguity.
For the first part, specialized gaze-estimation models appear to be approaching human-level precision \citep{icatcher_plus, hangaze2021, gazelle}, although our results suggest that their apparent success may be overestimated.
Their relative success typically relies on supervised learning with explicit gaze target annotations.
However, how to address the second part regarding skill integration remains unclear for these approaches.
Recent advances in Vision-Language Models \citep[VLMs;][\textit{inter alia}]{gemini0, llava} offer promising hosts for skill integration and domain generality.
These large-scale computational artifacts sometimes exhibit abilities that were not explicitly trained and can apply them across tasks and modalities.
Due to the lack of explicit training (in terms of gaze-specific training objectives), they might not have the same level of gaze inference competence.
Still, they remain potential candidates for solving the second challenge in principle, while specialized vision models are not powerful candidates for it.
This makes it necessary to understand how well VLMs can accurately infer gaze targets in zero-shot settings and produce insights that future improvement endeavors, such as finetuning, can build upon.
This study aims to address this empirical question and recommend potential improvement methods based on observed patterns.

\begin{figure}
  \includegraphics[width=\linewidth]{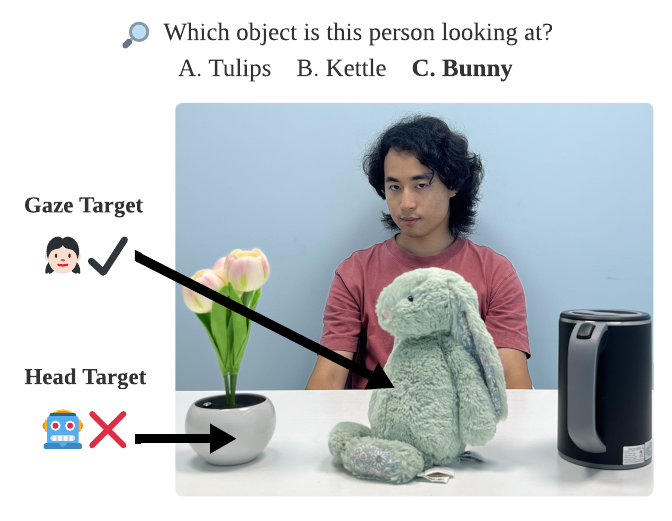}
  \caption{Example stimulus before size downscaling.}
  \label{fig: teaser}
\end{figure}

To this end, we formulated a task where VLMs are presented with an image and a multiple-choice question, illustrated in Fig.~\ref{fig: teaser}.
Each image depicts a single gazer seated at a table with two to four objects, looking at exactly one of them, with no other people present. 
VLMs are asked which object the person is looking at, with answer choices corresponding to the object names.

To arrive at targeted improvement suggestions, we go beyond benchmarking and characterize the gaze inference skill.
Although there is prior work that benchmarks VLMs' zero-shot gaze target inference \citep{vlmGaze2024}, only characterization can potentially yield recommendations for improvement.
This theoretical superiority comes with practical difficulty.
For example, blindly analyzing intermediate reasoning traits does not necessarily offer understanding \citep{cot1, cot2}.
One solution prevalent in cognitive sciences is a combination of piloting, systematically controlling variables, pre-determining hypotheses through pre-registration, and conducting statistical analysis.
By observing how manipulated factors affect behavior, we can constrain plausible theories about VLMs' underlying inferences by ruling out accounts that cannot explain the results.
Specifically, we control the number of objects on the table, the photo-shooting angle, the proximity \textit{between} objects, which object the gazer looks at, and, in some conditions, the gazer's head orientation.

As a toy example of constraining hypotheses, if the performance is lower when objects are closer to each other, then we at least know that VLMs are using the visual information in a task-relevant way and probably not bottlenecked by their (in)ability to locate eyes.
This kind of characterization is important because it can lower the likelihood of theories such as the Approximate Retrieval account \citep{Kambhampati_2024}, which posits that language model performance depends on a test stimulus's similarity to training data and predicts little correlation between performance and the proximity between objects when all other scene elements remain constant (as in Fig.~\ref{fig: stimuli_example} Subfigure b).

\begin{figure}
  \includegraphics[width=\linewidth]{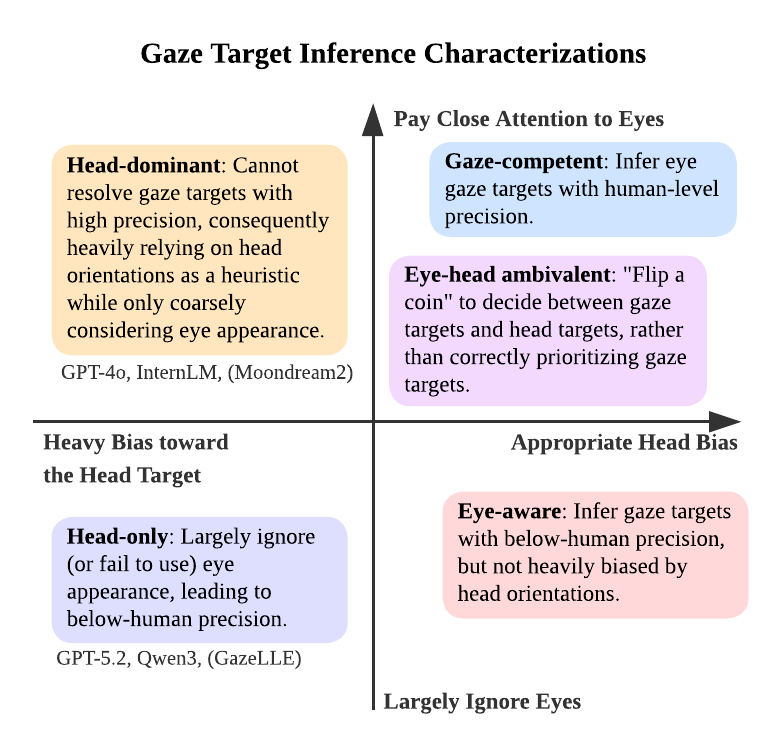}
  \caption{Vision-Language Models and baselines (bracketed) grouped by their inferred gaze target inference behavior. We found consistent head-orientation biases.}
  \label{fig: space}
\end{figure}

The main contribution of this work is a series of tests that allow us to sort any computational method into one of five hypothesized categories, as illustrated in Fig.~\ref{fig: space}, by presenting 10 repetitions of 1360 high-quality stimuli to each VLM.
Characterization categories are important because different mechanisms require different methods to improve.
The experiment design, stimuli, VLM sample size, power analysis script, and statistical analysis plan of the main study were pre-registered.
The main study is driven by a pre-pilot and a pilot that evaluated 111 VLMs on 900 pilot stimuli.
The main study and the pilot stimuli do not overlap and are both publicly accessible.\footnote{The pre-registration is at Open Science Framework: \url{https://osf.io/tj2wn}}

\section{Experiment Settings}
\subsection{Choice of VLMs}
After testing 111 VLMs in the pre-pilot, we chose GPT-4o-2024-08-06, GPT-5.2 \citep{gpt52}, Qwen3-VL-30B-A3B-Instruct \citep{Qwen3-VL}, InternLM-XComposer2-vl-7b, and GLM-4.6V \citep{vteam2025glm45vglm41vthinkingversatilemultimodal} for the main study.
To contextualize the results, we additionally tested two methods: Moondream2 2025-01-09 \citep{moondream}, a hybrid VLM with a specialized gaze inference function that involves Monte Carlo Sampling, and GazeLLE \citep{gazelle}, a finetuned vision model with a frozen pretrained transformer-based encoder backbone.
See details at Appendix~\ref{subsection: choosingVLMs}.

\subsection{Task Formulation}
\label{subsection: formulation}
The gaze target inference task features commitments that make it an experimental proxy to, rather than a comprehensive benchmark of, the underlying gaze inference skill, stemming from a need to trade off between isolating the core capacity of interest and accurately reflecting performance in the wild.

At the stimuli level, they are novel (i.e., photos that likely few in the training data resemble, at least not superficially), focused (e.g., no body orientation inference is needed; only a single person is present; and almost no background distractors), but still real-world photos (rather than synthetic images that may not share the same statistical regularities like shade and light with reality).
Firstly, the novelty feature mitigates the confounding factor of perceptual similarity to the training data.
Secondly, focused stimuli are easier to solve, such that the effect of manipulated variables is stronger.
Only when VLMs can solve these cases, more complex scenarios become interesting to test on.
Lastly, because deep-learning-based methods are good at discovering statistical correlations not visible to human eyes, real-world photos serve as a better starting point than synthetic pictures.

This collection of images does not and is not meant to comprehensively resemble the rich visual experience that people will perceive in their daily lives.
These artificially-constrained stimuli serve the end of isolating core capacity, rather than a measurement of VLM's gaze inference in a wide range of cases ("in the wild").
Performance in the wild involves not just the core capacity of interest, but also other factors like context, commonsense knowledge (e.g., people tend to look at things in their hands), and heuristics (e.g., focusing on body and head orientation without analyzing eye appearance).
We take the isolation of core capacity as the first step in an evaluation process.
Imagine an agent that excels in the wild but performs poorly when tested on the core capacity of interest with no peripheral information available.
It likely does not possess that capacity in the first place and will tend to be brittle in the edge cases.
Conversely, if VLMs are equipped with a robust core capacity of gaze inference, they will likely excel in the wild, due to their advantage of general knowledge.
If this first step indicates a robust capacity in simple cases with few confounders, in-the-wild evaluation will be better supported.
Indeed, many in-the-wild evaluations involve internet images where a head-orientation heuristic is often sufficient.
To pressure test VLMs, we isolate the core capacity by using stimuli that necessitate fine-grained analysis of eye appearance.

At the task level, three more commitments are made.
Firstly, we use a multiple-choice question format, rather than an open-ended, free-naming question without options or asking for coordinates.
The choices are names of objects on the table, in a randomized order for each trial.
This format avoids asking for coordinates as outputs, as in most gaze inference works \citep{icatcher_plus, hangaze2021, gazelle, vlmGaze2024}, which unnecessarily penalizes a correct answer on the sole basis of being different from the ``ground truth'' coordinate points decided by benchmark labelers, even if both coordinates are on the same correct object.
Secondly, we ask VLMs not to report intermediate reasoning and to output the option letter directly.
This instruction does not inherently prevent VLMs from free responding, as they might still choose to be verbose and reason.
Our evaluation pipeline is compatible with reasoning models by mapping free responses onto options (See Subsection~\ref{subsection: pipeline} for details).
Lastly, we instruct VLMs that they cannot refuse and ask them to make their best guess.
Because this task is particularly challenging for VLMs, some VLMs tend to refuse to choose an answer and simply give up.
Based on pilot results, asking them to make a best guess does not lead to performance change ($p > 0.23$ using a two-proportion test) while potentially bringing the benefit of a higher response rate.
These task-level commitments are reflected in the main study prompts, listed in the Appendix~\ref{table: prompts}.

\subsection{Characterizing VLM Gaze Inference Through Tests}
\label{subsection: hypotheses}
Controlling where the gazer looks and how they orient their heads allows the investigation of how VLMs integrate eye appearance and head orientation cues when inferring gaze target.
Head orientation is an informative heuristic, particularly when the gazer's eyes are not clearly visible.
Nonetheless, to excel on our stimulus set, VLMs need to weigh less on the head orientation cue and instead base their judgments largely on how the gazer's eyes look.
This reflects the idea that a full understanding of gaze requires weighing head orientation and eye appearance cues based on the situation.

\begin{figure}[!t]
  \includegraphics[width=\linewidth]{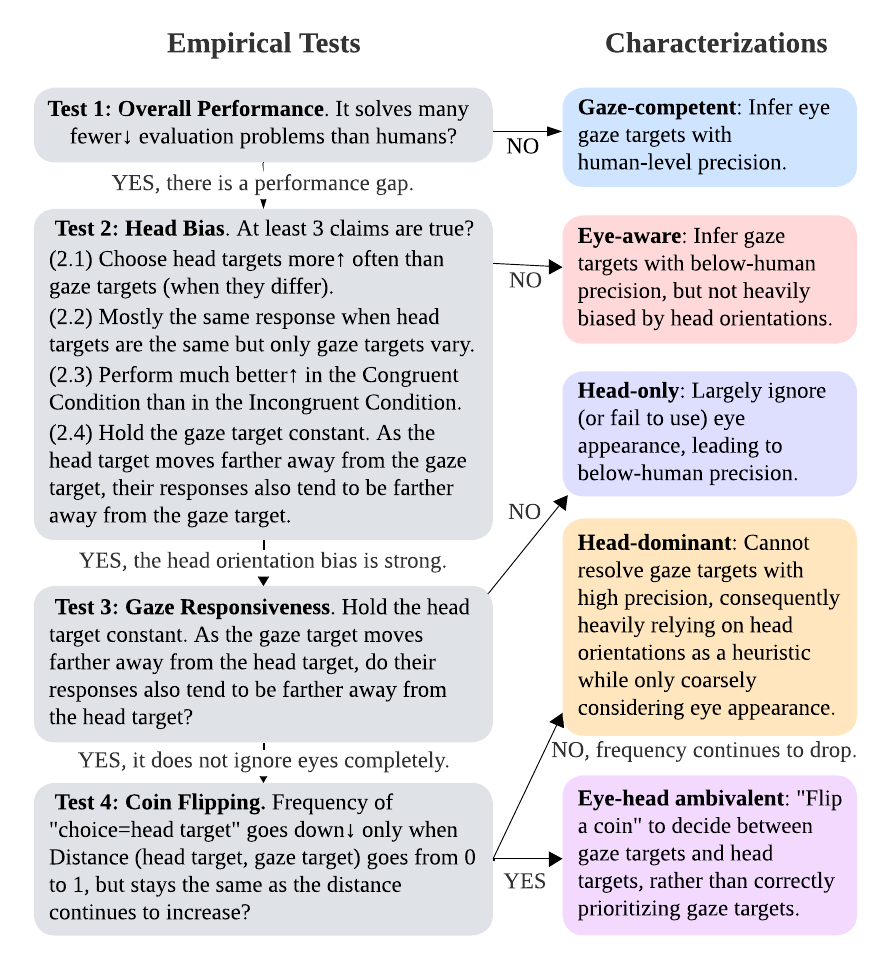}
  \includegraphics[width=\linewidth]{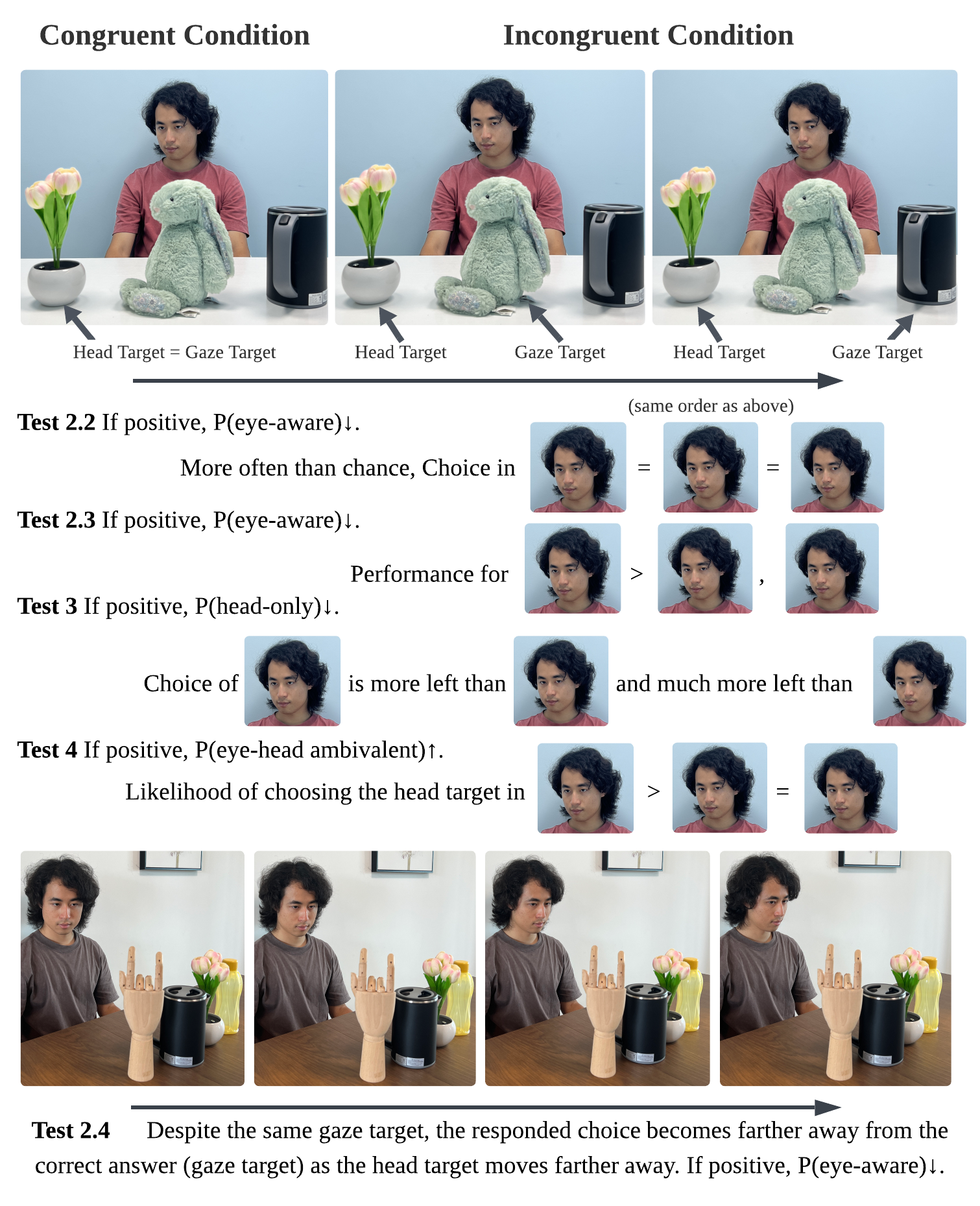}
  \caption{The first example involves holding \texttt{HeadTarget} constant when varying \texttt{GazeTarget}, while the second example is the opposite.}
  \label{fig: decisiontree}
\end{figure}

We propose to use a series of tests to identify what kind of cue weighing VLMs implement, summarized and illustrated in Fig~\ref{fig: decisiontree}.
The exact content of each test is described in Section~\ref{section: tests}.
The Overall-Performance Test (Test 1) clarifies whether VLMs infer gaze targets with human-level precision, as human gaze inference serves as a standard to compare against. 
If there is a performance gap between VLMs and humans, the next question is whether this difference is driven by VLMs' over-reliance on head orientation cues.
The Head-Bias Test (Test 2) consists of 4 sub-tests.
A positive result of each indicates a manifestation of the head orientation bias.
If there are at least 3 positive results from these 4 tests, then we can be confident about the VLM's over-reliance on head orientation.

If there is a heavy head bias, it is possible that the VLM effectively ignores eyeball appearance.
The Gaze Responsiveness Test (Test 3) examines this possibility by asking if VLMs are sensitive to which object is the gaze target at all.
If they do, their responses should be affected by the degree to which the gaze target is located far from the head target.
If so, there are still two possibilities.
VLMs might know which is the gaze target, but only fail to link this back to answer the question about "looking at" by prioritizing gaze targets (when available) above head targets.
VLMs sometimes respond with the head target, sometimes with the gaze target, as if they are "flipping a coin" to decide between the two.
Alternatively, they can only coarsely consider eye appearance, leading them to loosely select choices that lie between the gaze target and the head target.
These two possibilities are distinguished by the Coin Flipping Test (Test 4), which asks whether the number of objects located between the gaze target and the head target matters.

\subsection{VLM Evaluation Procedure}
\label{subsection: pipeline}
Each presentation to a VLM includes a visual stimulus and a textual prompt.
There are 1360 photos in the main study stimulus set. 
Each VLM is presented with each photo 16 times, corresponding to one of the 16 prompt templates, randomly sampled without replacement.
The prompt includes a multiple-choice question, formatting instructions, and potentially some additional hints.
The 16 prompts are listed in the Appendix~\ref{table: prompts}.
This makes 21760 (1360 times 16) trials for each VLM.
We justify the sample size by performing a priori sensitivity power analysis.
For an alpha of 0.05 and a power of 80\%, this sample size can detect what we found about proximity and photo-shooting angles in the pre-pilot.

To allow VLMs to freely respond within the multiple-choice question setup, we follow the pipeline developed by \citet{duan2024vlmevalkit}.
Parameters like the response decoding temperature and the maximal number of tokens are set to the ones recommended by each corresponding VLM provider.
After VLMs had made their full response, we matched their response with their options (A, B, C, or D).
This is done by first some manually defined matching templates, followed by semantic matching using a large language model judge (Meta-Llama-3.1-70B-Instruct) if template matching did not resolve the category, and then followed by authors' manual review if still not resolved.
The matching templates and the semantic matching process were hand-examined by sampling to ensure quality.
Responses that were truly not recognizable to human eyes are counted as incorrect responses, since they are often nonsensical responses when the problem is too difficult for the VLMs.
There are only 3 such instances in the main study, all of which are from the InternLM claiming that the gazer looks at the camera.
GPT-5.2 did not engage this procedure at all, perfectly following the instruction of only outputting the options.

\subsection{Manipulated and Dependent Variables}
\label{subsection: curation}
There are 3 conditions for how head orientation and ground truth interact.
In the Natural Condition, the gazer oriented his head however he wanted as long as he looked at the instructed eye gaze target comfortably. 
The pilot stimuli would have fallen under this condition. 
In the Congruent Condition, the gazer oriented his head towards an instructed target while also looking into the same target. 
The gazer adjusted his head orientation until he felt confident that his eye gaze and head orientation agreed.
Although self-perception might not be perfectly accurate, its quality is good enough for this experiment.
In the Incongruent Condition, the gazer kept his head still and moved his eyes to look at other objects on the table one by one, making the eye gaze target different from the head orientation target. 
Therefore, when switching from the Congruent Condition to the Incongruent Condition, and when switching within the Incongruent Condition, the only thing that differed was the eye appearance, and the head orientation remained constant, making the basis for Tests 3 and 4. 
To avoid carry-over effects, the gazer looked at the ceiling before every change of head orientations in all conditions.

\begin{figure}
  \centering
  \includegraphics[width=\linewidth]{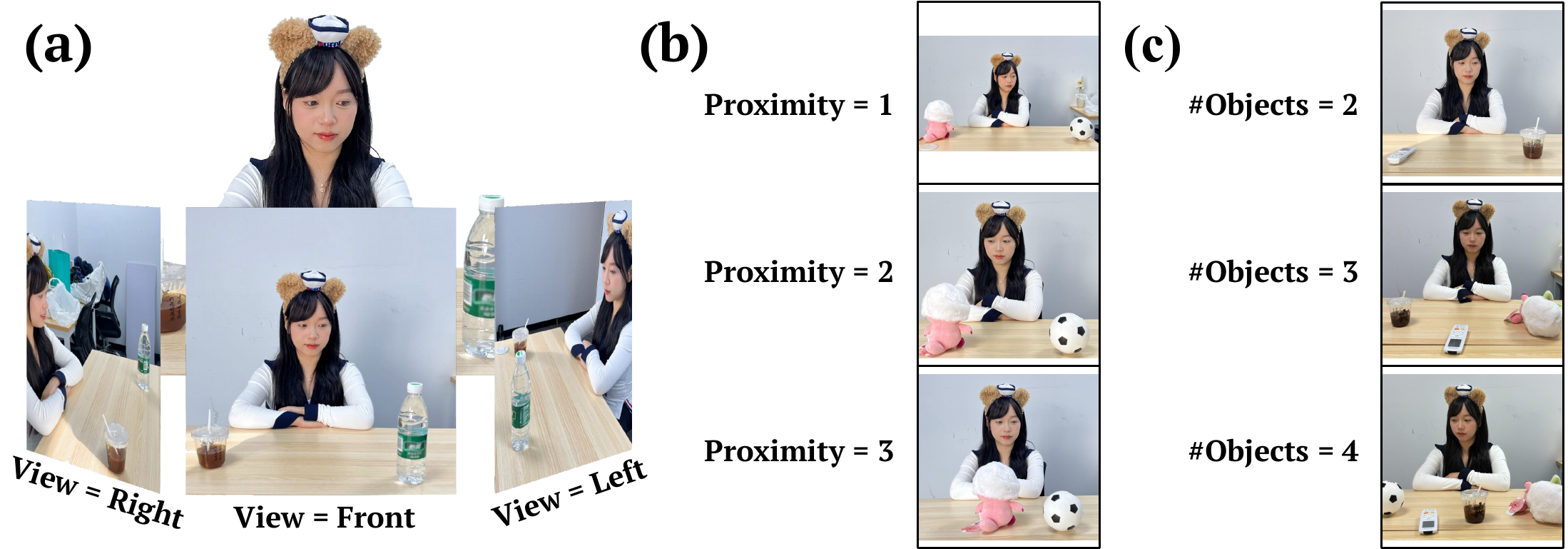}
  \caption{Manipulation of \texttt{View} (left/right/front), \texttt{Proximity} (1, 2, or 3), \texttt{\#Objects} (2-4), and \texttt{Objects} (13 combinations of 11 distinct items) across 1360 main study stimuli and 900 pilot stimuli. Stimuli in Subfigure (c) have a \texttt{Proximity} value of 2. Those shown are pilot stimuli, while main study stimuli follow the same logic but additionally control \texttt{HeadTarget} (not shown here).}
  \label{fig: stimuli_example}
\end{figure}

In summary, along with those variables briefly introduced earlier, the manipulated variables are:
\begin{itemize}
\setlength\itemsep{-3pt}
\item \texttt{\#Objects}: The number of objects on the table, from 2 to 4. A fixed effect.
\item \texttt{Objects}: The specific combination of objects on the table. There are a total of 11 distinct objects in the main study stimuli. There are 13 combinations, which are made up of 2 to 4 samples from these objects. A random effect of 13 levels.
\item \texttt{View}: Three possible values. Two views show the gazer’s left and right profile (\texttt{View=left/right}), and one shows a frontal view directly facing the gazer (\texttt{View=front}). A fixed effect.
\item \texttt{Proximity}: The distance between Objects (between each other, not to the gazer). Integer 1, 2, and 3, where a value of 3 represents the highest relative proximity (i.e., smallest distance). A fixed effect.
\item \texttt{PromptID}: Indicates which prompt template is presented. A random effect of 16 levels.
\item \texttt{StimulusID}: A unique identifier of the stimulus. A random effect of 1360 levels.
\item \texttt{GazeTarget}: The object the gazer looks at—the correct answer.
\item \texttt{HeadTarget}: The object the gazer's head orients to. This variable does not exist in the Natural Condition, in which the gazer oriented his head in a way he felt comfortable and natural.
\end{itemize}

Dependent variables:
\begin{itemize}
\setlength\itemsep{-3pt}
\item \texttt{Choice}: The response for each trial. The object that VLMs pick as the gaze target.
\item \texttt{Wrongness}: A magnitude of error, as a proportion of the maximum possible error. Computed by dividing the distance between \texttt{Choice} and \texttt{GazeTarget} (between 0 and 3) by the maximum possible distance conditioned on \texttt{GazeTarget} of this trial. Here, the distance is one plus the number of objects between them on the table. 
For example, if \texttt{\#Objects}=4, with the correct answer being the second from left, but a VLM chose the third from left, then \texttt{Wrongness}=1/2 and $\mathbb{E}$(\texttt{Wrongness})=1/2.
For any given Proximity and \#Objects, wrongness indicates the relative angular distance between the correct answer and chosen answer, with the gazer as the anchor point.
It reflects the precision of gaze target inference to a finer grain size than accuracy, serving as the default measure of performance. 
\item \texttt{DistEyeHead}: The distance between the \texttt{GazeTarget} and the HeadTarget on the table. Only for the Congruent and Incongruent Conditions (where \texttt{HeadTarget} exists).
\end{itemize}

\section{Performance Gap and Head-Bias}
\label{section: tests}
The investigation of Proximity and View effects involves the Natural and Congruent Condition, as the Incongruent condition is too different from pilot stimuli, where we first found the existence and the lack of these effects.
Except for Test 2.1, the rest of the tests involve the Congruent and the Incongruent conditions, because they are when the head orientation is controlled.

As \citet{continuous} suggested, we treated \texttt{\#Objects} and \texttt{Proximity} as continuous variables because they represent an underlying scale.
We mean-centered them for all tests.
\texttt{DistEyeHead} is similarly treated in all but the Coin Flipping Test, in which it is treated categorically to distinguish Head-dominant and Eye-head ambivalent.
For most tests, we fit multiple generalized linear mixed-effect models (GLMMs) and selected the most suitable one using a top-down model selection strategy. 
We began by first exploring the random effect structure, while fixing the random effect maximally complex.
The most complex random effect structure that can lead to converged models was used. 
Then, for better interpretability, we examined whether a no-interaction model is sufficient, using a Likelihood Ratio Test.
If yes, no-interaction models were used for coefficient reporting, while the complex ones were still used for visualization and marginal effect estimations.

\subsection{A Performance Gap Between VLMs and Humans (Test 1)}
We first tested 111 VLMs in the pre-pilot and 59 human participants in the pilot.
Each VLM saw the whole pilot stimulus pool of size 900, while each participant saw 45 questions sampled from the pool (see Appendix~\ref{subsection: human_survey} for human response collection details).
On the one hand, human participants can successfully infer the gaze target in 89\% of the questions.
On the other hand, the accuracy of most VLMs is very close to the expected accuracy of a machine that randomly selects a valid option in the multiple-choice question ($\mathbb{E}(\text{Accuracy})\approx42\%$).
Indeed, a two-sided test for proportions, assuming a normal approximation to the binomial distribution (i.e., a z-test), revealed that 94 of the 111 VLMs in the pre-pilot failed to perform significantly better than chance ($\alpha=.05$).
The pre-pilot performance summary is in Appendix~\ref{subsection: ranking}.

The same substantial performance gap persists for main study VLMs and even extends to baselines (Moondream2 and GazeLLE) when tested on the main study stimuli, Natural Condition.
The results are summarized in Table~\ref{table: accuracy}.
Note that both GazeLLE and Moondream2 (Hybrid) achieved human-level performance on GazeFollow \citep{gazefollow}, a benchmark that features in-the-wild stimuli from the internet.

\begin{table}[]
\resizebox{\columnwidth}{!}{%
\begin{tabular}{@{}llll@{}}
\toprule
Accuracy              & 2 Objs & 3 Objs & 4 Objs \\ \midrule
Humans (n=59)*        & 94\%   & 88\%   & 76\%   \\ \midrule
GazeLLE-DinoV2-VitL14 & 78\%   & 67\%   & 47\%   \\
Moondream2 (Hybrid)   & 78\%   & 58\%   & 41\%   \\ \midrule
GPT-5.2               & 64\%   & 46\%   & 31\%   \\
GPT-4o-20240806       & 65\%   & 41\%   & 30\%   \\
GPT-4o-20240806*      & 58\%   & 43\%   & 33\%   \\
Qwen3-VL-30B          & 59\%   & 39\%   & 28\%   \\
Qwen2.5-VL-72B*       & 54\%   & 40\%   & 30\%   \\
GLM-4.6V-Flash        & 62\%   & 43\%   & 30\%   \\
GLM-4V-9B*            & 54\%   & 42\%   & 33\%   \\
InternLM-XComposer2-VL-7B              & 64\%   & 43\%   & 29\%   \\
InternLM-XComposer2-VL-7B*             & 54\%   & 45\%   & 31\%   \\ \midrule
Guessing Baseline     & 50\%   & 33\%   & 25\%  
\end{tabular}%
}
\caption{Natural condition results. Moondream2 results were obtained by calling its specialized gaze function. (*) These results were extracted from the pilot study, which is roughly equally hard. }
\label{table: accuracy}
\end{table}

\subsection{Proximity and View Effects}
\begin{figure}[h]
\centering
\includegraphics[alt={}, width=\columnwidth]{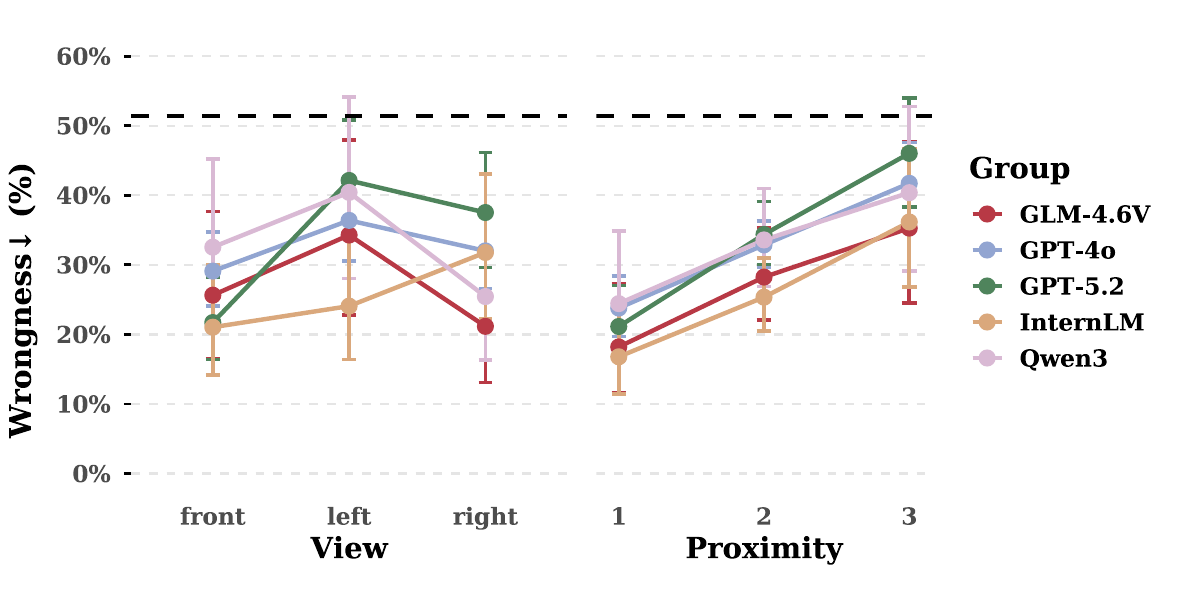}
\caption{The estimated marginal means of Wrongness. The lower, the better. The random-guessing baseline is indicated by dashed lines. Statistical models are separately fitted for each VLM. }
\label{fig: effects}
\end{figure}

Fig.~\ref{fig: effects} depicts an intuitive summary of these two effects, while trends in the log-odds space (where effects are detected) are in Appendix~\ref{subsection: appendix_modeling}.
The Proximity effect posits that as \texttt{Proximity} increases (i.e., \texttt{Objects} become closer to each other), performance decreases, manifesting as an increase in \texttt{Wrongness}.
The View effect posits that \texttt{Wrongness} is higher when viewing from the sides (left/right) compared to the front.
While these effects can be defined using accuracy, \texttt{Wrongness} is a finer-grain measure.
The presence of a Proximity effect indicates that VLMs utilize visual information in a task-relevant way.
A lack of the View effect indicates low sensitivity to eye appearance, as eye appearance differences are smaller when viewing from the sides.
A strong View effect, on the other hand, can be due to sensitivity to eyes, but can also be due to lower competence in three-dimensional direction tracking from a side view.

To test the reliability of these findings in VLMs, we fit a binomial GLMM:
\[
\begin{aligned}
\mathrm{logit} (\mathrm{Wrongness})
\sim 
\beta_0 + \mathrm{logit}(\mathbb{E}(\mathrm{Wrongness})) \\
+ \mathrm{View} \times \mathrm{Proximity} \times  \mathrm{{\#Objects}} \times \mathrm{Condition} \\
+ \gamma_\mathrm{StimulusID} + \gamma_\mathrm{PromptID} + \gamma_\mathrm{Objects},
\end{aligned}
\]
where we treated \texttt{Wrongness} as a binomial proportion, such that a \texttt{Wrongness} of $\frac12$ is treated as one ``success'' and one ``failure''.
The log-odds function $\text{logit}:= p \mapsto\ln ({p}/{1-p})$ is the inverse of the logistic function $\sigma(x) = 1/\big[{1+\exp(-x)}\big]$.
We included a trial-specific offset term such that the fixed and random effects need to account for the variance of the log-odds of the ratio between the observed and the trial-specific expected value of \texttt{Wrongness}.
See Appendix~\ref{subsection: appendix_modeling} for details of modeling.

We then estimated the marginal effects to determine the presence of Proximity and View effects. 
During pre-pilot, GLM-4V and Gemini showed the Proximity effect (p=0.004, 0.001), but not Qwen2.5 (p=0.150). 
None of them shows the View effect, but humans do (p<0.001).
In the main study, GPT-5.2 showed Proximity and View effects (both $p<1e-4$).
The existence of the Proximity effect and the lack of the View effect is found in GPT-4o, InternLM (both as in the pilot), GLM-4.6V, and Qwen3.

\subsection{Is There a Head Orientation Bias? (Tests 2, 3, and 4)}
See Table~\ref{table: tests} for a summary of test results.
Test 2.1 asks whether VLMs chose \texttt{HeadTarget} More Often than \texttt{GazeTarget}, for the Incongruent Condition with \texttt{\#Objects}>2.
Qwen3 and GPT-5.2 are positive, indicating an extremely strong head bias.
Test 2.2 asks whether VLMs responded with the same \texttt{Choice} when only \texttt{GazeTarget} varied while all other scene elements except eye appearance remained constant.
For all VLMs, the number of scenes where this is true is significantly higher than chance.
It is also true for all VLMs that their performance is higher when \texttt{HeadTarget} coincides with \texttt{GazeTarget} than when they differ (Test 2.3), and that Distance(\texttt{Choice}, \texttt{GazeTarget}) Increases as Distance(\texttt{HeadTarget}, \texttt{GazeTarget}) Increases (Test 2.4).
For all VLMs, the effects in Test 2.4 are consistently stronger for stimuli with smaller \texttt{Proximity}, potentially reflecting a heavier reliance on head orientation in more difficult cases.
Gaze Responsiveness Test (Test 3) similarly asks whether Distance(\texttt{HeadTarget}, \texttt{Choice}) Positively Correlates with Distance(\texttt{HeadTarget}, \texttt{GazeTarget}).
Replicating the pattern in Test 2.1, Qwen3 and GPT-5.2 again showed a head bias so strong that it effectively ignored eye appearance.
Lastly, the Coin Flipping Test (Test 4) aims to rule out the unlikely possibility that a VLM can infer gaze targets accurately but only fails to understand that the eye appearance cue precedes head orientation (Eye-head ambivalent).

\begin{table}[]
\resizebox{\columnwidth}{!}{%
\begin{tabular}{@{}llllll@{}}
\toprule
               & View & Test 2.1 & Test 3 & Test 4                                  & Types   \\ \midrule
(GazeLLE)        & -    & +        & -  & Neither                             & Head-only  \\
(Moondream2)     & -    & +        & +  & Head-dominant               & Head-dominant  \\
GPT-5.2        & +    & +        & -  & Neither                             & Head-only  \\
GPT-4o         & -    & -        & +  & Head-dominant & Head-dominant  \\
Qwen3-VL-30B   & -    & +        & -  & Neither                             & Head-only  \\
GLM-4.6V-Flash & -    & -        & +  & Neither                             & Unclear \\
InternLM       & -    & -        & +  & Head-dominant               & Head-dominant  \\ \bottomrule
\end{tabular}%
}
\caption{"+"/"-" indicates a positive/negative test result. Moondream2 results were obtained by calling its specialized gaze function, while GazeLLE is a finetuned model. Test 2.2, 2.3, and 2.4 results are all positive. All exhibited the Proximity effect.}
\label{table: tests}
\end{table}

\section{Why is There a Head Bias?}
We need to rule out alternative explanations before we can attribute the performance gap to head orientation bias and eye appearance processing deficit.
\begin{itemize}
\item The ability to map names onto objects in the scene. VLMs likely possess this ability, because Moondream3-preview-base (a 9B VLM used for annotation to evaluate GazeLLE and Moondream2) rarely finds a wrong coordinate of a named object (only 25/8000 of the time for the full-resolution version of the main study and pilot stimuli, and it should not be significantly harder after size downscaling). This is further supported by the marginal improvement from providing object names in the image from left to right in the prompts (only GPT-5.2 and GPT-4o showed a significant improvement by 2\% and 1.5\% on Natural and Congruent Conditions, with two-sided paired t-test, $p=0.002, 0.01$, respectively).
\item Task understanding. There is no significant performance difference whether VLMs were given explicit instruction on how to solve the task or not, for all VLMs but InternLM ($p=0.03, \text{accuracy difference}=0.8\%$, two-sided paired t-test). See the list of prompts in Appendix~\ref{table: prompts}.
\item Evaluation procedure. GPT-5.2 output letters only and did not engage the option matching procedure at all, such that the performance gap cannot be explained away by the option matching design.
\end{itemize}

It appears that there is a head orientation bias and/or eye appearance processing deficit, but why?

\begin{itemize}
\item Resolution cannot be the only reason. Human participants who saw the same images of the same resolution (448x448) were able to solve most of the stimuli.
In contrast, GPT-5.2 solves around 39\% of a random subset of 100 main study stimuli with this resolution ($95\%\ CI=[0.366, 0.414]$).
Resolutions of 896x896 ($M_{accuracy}=0.34, 95\%\ CI=[0.247, 0.433]$) and 1024x1024 ($M_{accuracy}=0.39, 95\%\ CI=[0.294, 0.486]$) do not improve accuracy.

\item Attentional skill cannot be the only reason. VLMs might find it hard to attend to the small eye regions because of their limited adaptive attentional skills.
This still cannot be the sole reason, because they also showed sub-human precision in the Congruent Condition, where the head orientation alone predicts the correct answer, while the head occupies a much larger space.
Failure to attend to the feature of head orientation among many features of the head can be a reason.
\item Patch-by-patch processing cannot be the sole reason. VLMs convert an image into patches, which are in turn converted into tokens. There might be limited information these tokens can contain, given the context length limit. This is likely a reason, but it cannot be the sole reason, as GazeLLE also showed head orientation bias and eye appearance processing deficit.
\item Increasing parameter size is not helpful. Later VLMs are not better. See Appendix~\ref{subsection: scaling}.
\end{itemize}

We think the most likely and major cause is the training data.
Images that contain faces and clear eye details are sparse in VLM training data.
In addition, humans, as deeply social and cultural creatures both evolutionarily and individually, have a great need to interact with other humans and recognize their gaze.
Training signals (or selection pressure signals) of gaze inference come from not just the physical domain but also the social domain.
These statistical correlations between the visual world and the social world prevalent in the human social world might be less reflected on the internet.
Training data that mostly consists of internet content might teach VLMs very little about the interplay of different visual cues and where someone looks.
To test this hypothesis that head-bias is mainly driven by data, we provide a proof-of-concept experiment by finetuning GazeLLE.

\section{Mitigating Head-Bias: A Proof-of-Concept Experiment}

The preceding analysis suggests that the observed head-orientation bias may arise from the distribution of training data, which likely overrepresents cases where head orientation aligns with gaze direction and underrepresents counterexamples requiring fine-grained eye-based inference.
Another approach to further support this claim is conducting a proof-of-concept experiment by fine-tuning GazeLLE on our stimulus set. 
The combined stimulus set from the pilot and main study (of size $900+1360$) is randomly split into training, validation, and test sets with a ratio of $7:1:2$. Following the original GazeLLE training setup, the training objective is to minimize the Euclidean distance between the predicted gaze point and the ground-truth target location of the gaze referent object. We fine-tune the model for $50$ epochs using a cosine annealing learning rate schedule starting from $1e-3$ to $1e-6$. The training accuracy plateaus and converges to approximately $98.10\%$, showing that the model fits the training data effectively.

\begin{table}[t]
\resizebox{\columnwidth}{!}{%
\begin{tabular}{@{}lcccc@{}}
\toprule
               & Pilot & Congruent & Incongruent & Natural \\ \midrule
GazeLLE (ViT-B/14) & 63.80 & 62.82 & 15.65 & 65.22 \\
Fine-tuned         & 85.77 $\pm$ 1.40 & 70.00 $\pm$ 5.01 & 34.15 $\pm$ 1.76 & 76.81 $\pm$ 2.90 \\
\bottomrule
\end{tabular}%
}
\caption{Test Accuracy (\%) before and after finetuning the pretrained GazeLLE model (mean $\pm$ std over 5 runs of finetuning).}
\label{tab:gazelle_finetune_results}
\end{table}

Fine-tuning leads to substantial improvements across all conditions. 
We aggregate results over five independent finetuning runs with different random seeds using accuracy as the primary metric.
Table~\ref{tab:gazelle_finetune_results} breaks down test set performance by pilot stimuli, congruent, incongruent, and natural conditions. 
Performance on pilot and natural conditions improves consistently, with gains of approximately $21.97\%$ and $11.59\%$, respectively, suggesting broader improvements in general gaze inference. The improvement in the congruent condition is most modest (from $62.82\%$ to $70.00\%$, with a large standard deviation), indicating that the pretrained model already performs reasonably well when head orientation aligns with gaze direction. The most impressive gain is observed in the incongruent condition, where accuracy increases from $15.65\%$ to $34.15\%$, consistent with the prediction that fine-tuning enables the model to rely less on head orientation cues. 

\begin{figure}[t]
\centering
\includegraphics[alt={}, width=\columnwidth]{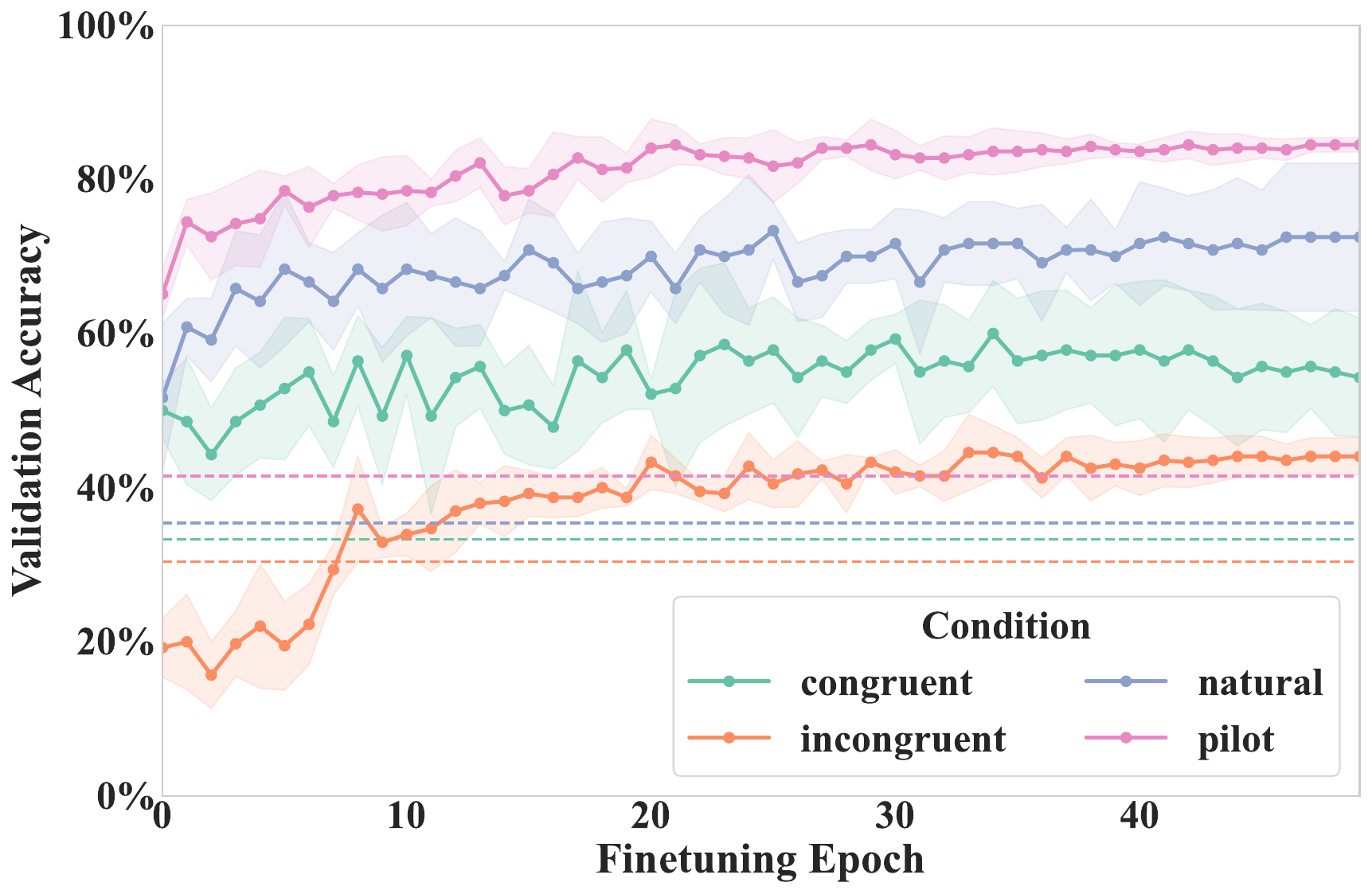}
\caption{Accuracy (\%) of GazeLLE variants on the validation set across finetuning epochs using our stimuli (7:1:2 split). The shaded bands correspond to $\pm$ one standard deviation around the mean over five independent runs of finetuning. Incongruent condition performance goes from below-chance to above-chance, indicating mitigation of head bias.}
\label{fig: gazelle_finetuned_across_seeds}
\end{figure}

The finetuning dynamics further support this interpretation. 
Figure~\ref{fig: gazelle_finetuned_across_seeds} shows validation accuracy across finetuning epochs.
Accuracy in the Incongruent condition increases rapidly during the first 10 epochs, whereas other conditions exhibit more gradual and stable improvements. This early-phase jump suggests that the model quickly adapts to the case where head orientation is misleading, reducing its reliance on shortcut heuristics. 
Despite these gains, performance in the incongruent condition remains the lowest, indicating that head-orientation bias is mitigated but not fully eliminated, potentially driven by the small stimulus set size. Detailed per-run validation dynamics are reported in Appendix~\ref{fig: gazelle_dynamics_per_run}.

\section{Conclusion}
Identifying likely causes of the performance gap allows targeted recommendations that might bridge this gap.
Incorporating training data that explicitly captures the interaction between body orientation, head orientation, and eye appearance, particularly in cases where these cues are misaligned, may be beneficial.
In addition, training examples that minimize reliance on contextual cues or common knowledge could encourage models to develop a more robust understanding of gaze based on visual evidence alone.
Once equipped with this capability, VLMs can be further trained to integrate gaze inference into downstream tasks across domains, enabling performance that more closely aligns with human behavior \citep{dnns_model_biology}. These include combining contextual information with visual reasoning, grounding verbal communication through gaze cues, inferring intentionality, and performing perspective-taking \citep{vpt_benchmark}.
Targeted training data where gaze cues, but not other factors, are necessary to make sense of the situation might be helpful for the acquisition of these higher-level skills.

By employing a controlled study to probe core capacity, we show that current VLMs are not yet able to reliably infer human gaze direction. More broadly, our work serves as a proof-of-concept for the use of hypothesis-driven evaluation and controlled behavioral probing, contributing to a growing body of research advocating for such methodologies in the study of vision-language models.

\section{Limitations}
The incongruent condition is a central condition for our study, but it might initially appear to lack ecological validity.
Gaze-target inference relies on integrating at least three cues: body orientation, head orientation, and eye appearance.
To understand how these cues interact when they do not align, we manipulated the degree of incongruence (\texttt{HeadTarget}).
For example, a DistEyeHead of 1 indicates looking at an object while the head faces an adjacent one. 
Because head-turning incurs a higher biomechanical cost than eye movement, minor incongruencies (\texttt{DistEyeHead} of 1) are likely both frequent and natural.
While higher degrees of incongruence are undeniably less common in real-world interactions, they are not entirely rare, as suggested by the existence of a phrase for it (“sidelong glances”). 
However, we currently lack empirical data quantifying the exact real-world frequencies of these specific cue misalignments, which constitutes a primary limitation of this work.

A second limitation is the lack of a diverse set of actors and actresses.
There are two actresses in our pilot and one actor in our main study.
Besides the severe resource constraint when it comes to human models, it can be an additional confounding variable if the gazer also varies when we systematically control those manipulated variables, making our effects underpowered and difficult to detect.
There are also varieties across gazers, particularly facial features like fake eyelashes and colored contact lenses.
Since we pre-registered the main study and do not see substantial performance differences between stimuli with these gazers, we believe that even if, in the unlikely case that all these stimuli happen to be extremely difficult for VLMs but not human participants, the performance gap between VLMs and humans cannot be solely attributed to this factor.
We do agree that having a more diverse set will solidify and contextualize our findings better.

\section{Ethical Considerations}
Gemini Pro was used for statistical advice and writing.
Claude Pro was used for coding.
We followed the Terms of Service for proprietary (API-Based) as well as licenses for openly distributed VLMs.
\paragraph{Societal Implications}
Our results have implications for understanding how VLMs might impact our society.
Importantly, the possibility of machines that can understand gaze targets should be taken seriously.
Users may not want machines to interpret their nonverbal cues without their explicit consent, for example, when they are inputting passwords on a keyboard. 
Users of VLMs with such an ability should be warned about their capabilities before using them.
Our results indicate that VLMs do not currently possess this skill, yet the possibility remains.
Further development of this ability in VLMs should be taken with great care.

\paragraph{Human Participants}
For human response collection, all methods were approved by the Johns Hopkins Institutional Review Board and were carried out per relevant guidelines and regulations. 
Human response is de-identified by only keeping the Prolific ID.
Our consent form informed that we will collect limited demographics and contact details (as required) for eligibility. 
Data are used only for research; publications use aggregate/de-identified results. 
De-identified data may be shared with qualified researchers/regulators and retained long term; identifiable records follow legal retention.
All participants are paid an hourly rate of 12 USD regardless of their residence.
 
\paragraph{The Human Actor and Actresses}
For stimuli collection, the one actor and two actresses granted us permissive, royalty-free rights to employ their visual materials.
To maximize reproducibility, the stimuli that contain their faces are publicly accessible.
While we are unable to remove their faces, which are the whole point of this study, the stimuli and annotation sheet do not contain identifiable textual information like names.
Actresses received financial compensation, while the actor agreed to volunteer.
We thank them for their contribution.

\section{Acknowledgment}
We thank \href{https://zillion.network}{Zillion Network Inc.} for providing the computation resources used in this work. Their optimized peak/off-peak scheduling, high-throughput storage infrastructure, and automated environment management enabled the cost-efficient and reliable execution of our experiments.
We also thank the \href{https://dash.nrp-nautilus.io/}{NRP Nautilus} cluster for generously providing the compute and storage support for the experiments in this paper.
The authors would also like to thank all anonymous reviewers for their valuable feedback.

\newpage
\bibliography{bib_ai, bib_psych, bib_ml}


\appendix

\clearpage
\section{Appendix}

\subsection{Choosing VLMs to Evaluate}
\label{subsection: choosingVLMs}
In the pre-pilot, we tested 111 VLMs. 
Given that most VLMs performed close to chance, it was only meaningful to analyze the most performant ones closely. 
Therefore, we ran a larger-scale pilot for 5 VLMs: GPT-4o-2024-08-06 \citep{gpt4o}, Gemini 1.5 Pro 002 \citep{gemini}, Qwen2.5-VL-72B-Instruct \citep{qwen2.5-VL}, InternLM-XComposer2-vl-7b \cite{internlm}, and GLM-4V-9B \citep{glm2024chatglm}.
They were five of the top seven most-performant VLMs in the pre-pilot, while InternLM-XComposer2d5-7b and OpenAI-o1 were not selected for redundancy and cost considerations, respectively. 
For the main study, we tested: GPT-4o-2024-08-06, GPT-5.2-2025-12-11 \citep{gpt52}, Qwen3-VL-30B-A3B-Instruct \citep{Qwen3-VL}, InternLM-XComposer2-vl-7b, and GLM-4.6V \citep{vteam2025glm45vglm41vthinkingversatilemultimodal}. 
They are mostly from the pilot or newer versions of those in the pilot, except for GPT-5.2, which is chosen to represent the closed-source frontier because Gemini-3-Pro-preview has too strict a rate limit. 
This way, they not only represent the new frontier but also connect with our pilot.

\subsection{The Pilot and Main Study Stimuli}
\label{subsection: more_stimulus}
\label{subsection: stimuli_dist}
\begin{figure}[H]
\centering
\includegraphics[alt={}, width=\linewidth]{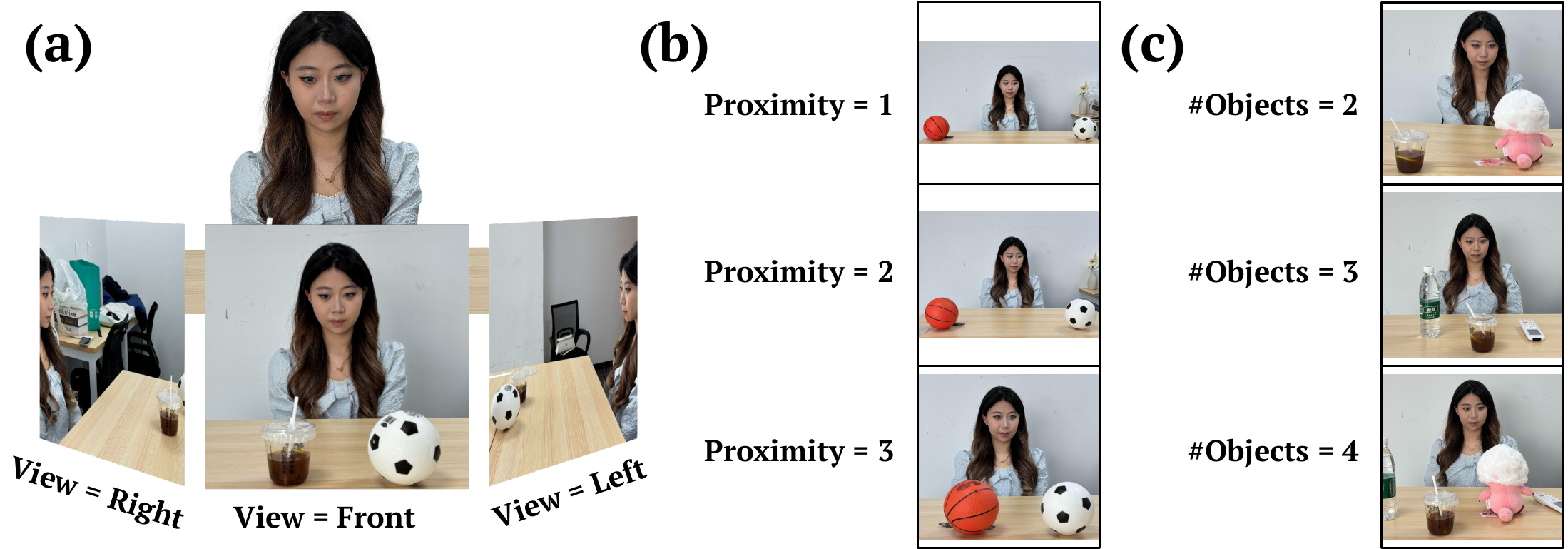}
\caption{Examples of stimuli for Actor Y with different \texttt{View}, \texttt{Proximity} and \texttt{\#Objects}.}
\label{fig: actor_y}
\end{figure}

The pilot stimuli feature actresses X and Y.
Cropping is used to maximize the proportion of space that relevant visual information occupies.
The main study stimulus features actor Z.
The camera is an iPhone 13 Pro with its default setup for square photo shooting. 
There are different facial features, such as fake eyelashes and colored contact lenses.
Examples of actress X stimuli are shown in Figure~\ref{fig: stimuli_example}.
Fig.~\ref{fig: actor_y} shows examples of actress Y.

\paragraph{Scene Construction}
All rooms were well-lit.
No catch light in the eyes.
A diverse set of objects (9 in the pilot stimuli and another 11 in the main study stimuli). 
Mixed-effect modeling reveals that objects as a random effect lead to singularity, so no difference in this dimension explains the outcome.
There are 3 environments: blue wall, white wall, and office (A.7.8). The blue-wall room has no background. The pilot stimuli were collected from the office, while the main study stimuli were collected from the first two. The white-wall room has a decorative wall canvas. 

\paragraph{Proximity Rubrics}
For \texttt{Proximity}, a value of one corresponds to putting objects farthest away from each other on the table, while a value of three means they are placed the closest possible, but not touching.
It represents the relative sparseness of objects, rather than absolute distance.

\paragraph{Stimulus Cleansing Procedure}
We manually examined every photo we took and dropped all photos taken in the middle of a blink.
All cases with occluded eyes can be approached by considering the head position, and there is no significant occlusion of parts of the objects.
Black padding was added when resizing photos to 448 by 448 pixels for pilot stimuli, while the main stimuli were already square.

\paragraph{Generalizability}
Note that sometimes the background is messy: this increases the credibility of our evaluation.
Messy backgrounds mainly appear when \texttt{View=right}, yet we do not observe significantly worse performance in the  \texttt{View=right} condition, so such a background introduces minimal confounding effect while adding realistic noise to the stimuli and increases diversity.
We thus expect our results to have reasonable generalizability.

\subsection{VLM Evaluation Details}
\label{subsection: prompts}
See Table.~\ref{table: prompt1}, \ref{table: prompt2}, and \ref{table: prompts} for the list of prompt templates we used for the pre-pilot, pilot, and main study, respectively.
The image tokens were inserted into the place marked by the placeholder in the template.
\texttt{\textbackslash n} indicates a change of line.

The encouragement of guessing when uncertain is a response to VLM refusal. 
In the pre-pilot, they were explicitly asked to guess when uncertain. 
In the pilot, some prompts only said that they cannot refuse to choose. 
No significant difference was found between runs with different prompts, so the problem does not lie in the explicit instruction of guessing. 
There is still a chance that the firm attitude failed some VLMs. 
Unfortunately, since this task is particularly challenging for VLMs, many VLMs will refuse to choose most of the time when the forced-choice instruction is absent. 
Allowing models to opt out would make it impossible to assess their performance comprehensively, as a 'strategic' model could achieve high accuracy simply by answering only the easiest questions. 
Our approach ensures we can characterize model behavior across the full range of difficulty, which is essential for our analysis.

\begin{table}[H]
\scriptsize
\centering

\caption{The full list of $12$ prompt templates in the pre-pilot, assuming that \texttt{\#Objects=3}.} 
\label{table: prompt1}
\vskip 0.12in
\begin{tabular}{|>{\centering\centering}p{0.3cm}|p{7cm}|}
\hline
ID & Prompt Template \\ \hline
1 & <image> What is this person looking at?\textbackslash n A. xxx\textbackslash n B. xxx\textbackslash n C. xxx\textbackslash n Please answer with the option's letter A, B, C directly. If you don't know, you still must choose one, so you might select randomly. You cannot refuse to choose. \\ \hline
2 & <image> Where is this person looking?\textbackslash n A. xxx\textbackslash n B. xxx\textbackslash n C. xxx\textbackslash n Please answer with the option's letter A, B, C directly. If you don't know, you still must choose one, so you might select randomly. You cannot refuse to choose. \\ \hline
3 & <image> Which object is this person looking at?\textbackslash n A. xxx\textbackslash n B. xxx\textbackslash n C. xxx\textbackslash n Please answer with the option's letter A, B, C directly. If you don't know, you still must choose one, so you might select randomly. You cannot refuse to choose. \\ \hline
4 & <image> What is she looking at?\textbackslash n A. xxx\textbackslash n B. xxx\textbackslash n C. xxx\textbackslash n Please answer with the option's letter A, B, C directly. If you don't know, you still must choose one, so you might select randomly. You cannot refuse to choose. \\ \hline
5 & <image> Where is she looking?\textbackslash n A. xxx\textbackslash n B. xxx\textbackslash n C. xxx\textbackslash n Please answer with the option's letter A, B, C directly. If you don't know, you still must choose one, so you might select randomly. You cannot refuse to choose. \\ \hline
6 & <image> Which object is she looking at?\textbackslash n A. xxx\textbackslash n B. xxx\textbackslash n C. xxx\textbackslash n Please answer with the option's letter A, B, C directly. If you don't know, you still must choose one, so you might select randomly. You cannot refuse to choose. \\ \hline
7 & <image> What is this person looking at in the image?\textbackslash n A. xxx\textbackslash n B. xxx\textbackslash n C. xxx\textbackslash n Please answer with the option's letter A, B, C directly. If you don't know, you still must choose one, so you might select randomly. You cannot refuse to choose. \\ \hline
8 & <image> Where is this person looking in the image?\textbackslash n A. xxx\textbackslash n B. xxx\textbackslash n C. xxx\textbackslash n Please answer with the option's letter A, B, C directly. If you don't know, you still must choose one, so you might select randomly. You cannot refuse to choose. \\ \hline
9 & <image> Which object is this person looking at in the image?\textbackslash n A. xxx\textbackslash n B. xxx\textbackslash n C. xxx\textbackslash n Please answer with the option's letter A, B, C directly. If you don't know, you still must choose one, so you might select randomly. You cannot refuse to choose. \\ \hline
10 & <image> What is she looking at in the image?\textbackslash n A. xxx\textbackslash n B. xxx\textbackslash n C. xxx\textbackslash n Please answer with the option's letter A, B, C directly. If you don't know, you still must choose one, so you might select randomly. You cannot refuse to choose. \\ \hline
11 & <image> Where is she looking in the image?\textbackslash n A. xxx\textbackslash n B. xxx\textbackslash n C. xxx\textbackslash n Please answer with the option's letter A, B, C directly. If you don't know, you still must choose one, so you might select randomly. You cannot refuse to choose. \\ \hline
12 & <image> Which object is she looking at in the image?\textbackslash n A. xxx\textbackslash n B. xxx\textbackslash n C. xxx\textbackslash n Please answer with the option's letter A, B, C directly. If you don't know, you still must choose one, so you might select randomly. You cannot refuse to choose. \\ \hline
\end{tabular}
\end{table}

\begin{table}[H]
\scriptsize
\centering

\caption{The full list of $12$ prompt templates in the pilot, assuming that \texttt{\#Objects=3}.} 
\label{table: prompt2}
\vskip 0.12in
\begin{tabular}{|>{\centering\arraybackslash}p{0.3cm}|p{7cm}|}
\hline
ID & Prompt Template \\ \hline
1 & <image> What is this person looking at?\textbackslash n A. xxx\textbackslash n B. xxx\textbackslash n C. xxx\textbackslash n Please answer with the option's letter A, B, C directly. If you don't know, you still must choose one, so make your best guess. \\ \hline
2 & <image> Where is this person looking?\textbackslash n A. xxx\textbackslash n B. xxx\textbackslash n C. xxx\textbackslash n Please answer with the option's letter A, B, C directly. If you don't know, you still must choose one, so you might select randomly. You cannot refuse to choose. \\ \hline
3 & <image> Which object is this person looking at?\textbackslash n A. xxx\textbackslash n B. xxx\textbackslash n C. xxx\textbackslash n Please answer with the option's letter A, B, C directly. You cannot refuse to choose. \\ \hline
4 & <image> What is she looking at?\textbackslash n A. xxx\textbackslash n B. xxx\textbackslash n C. xxx\textbackslash n Please answer with the option's letter A, B, C directly. There is no need to reason. If you don't know, you still must choose one, so make your best guess. \\ \hline
5 & <image> Where is she looking?\textbackslash n A. xxx\textbackslash n B. xxx\textbackslash n C. xxx\textbackslash n Please answer with the option's letter A, B, C directly. There is no need to reason. If you don't know, you still must choose one, so you might select randomly. You cannot refuse to choose. \\ \hline
6 & <image> Which object is she looking at?\textbackslash n A. xxx\textbackslash n B. xxx\textbackslash n C. xxx\textbackslash n Please answer with the option's letter A, B, C directly. There is no need to reason. You cannot refuse to choose. \\ \hline
7 & <image> What is this person looking at in the image?\textbackslash n A. xxx\textbackslash n B. xxx\textbackslash n C. xxx\textbackslash n Please answer with the option's letter A, B, C directly. If you don't know, you still must choose one, so make your best guess. \\ \hline
8 & <image> Where is this person looking in the image?\textbackslash n A. xxx\textbackslash n B. xxx\textbackslash n C. xxx\textbackslash n Please answer with the option's letter A, B, C directly. If you don't know, you still must choose one, so you might select randomly. You cannot refuse to choose. \\ \hline
9 & <image> Which object is this person looking at in the image?\textbackslash n A. xxx\textbackslash n B. xxx\textbackslash n C. xxx\textbackslash n Please answer with the option's letter A, B, C directly. You cannot refuse to choose. \\ \hline
10 & <image> What is she looking at in the image?\textbackslash n A. xxx\textbackslash n B. xxx\textbackslash n C. xxx\textbackslash n Please answer with the option's letter A, B, C directly. There is no need to reason. If you don't know, you still must choose one, so make your best guess. \\ \hline
11 & <image> Where is she looking in the image?\textbackslash n A. xxx\textbackslash n B. xxx\textbackslash n C. xxx\textbackslash n Please answer with the option's letter A, B, C directly. There is no need to reason. If you don't know, you still must choose one, so you might select randomly. You cannot refuse to choose. \\ \hline
12 & <image> Which object is she looking at in the image?\textbackslash n A. xxx\textbackslash n B. xxx\textbackslash n C. xxx\textbackslash n Please answer with the option's letter A, B, C directly. There is no need to reason. You cannot refuse to choose. \\ \hline
\end{tabular}
\end{table}

\begin{strip}
\begin{table}[H]
\scriptsize
\centering
\caption{The full list of $16$ prompt templates in Main Experiment, assuming that \texttt{\#Objects=3}.} 
\vskip 0.12in
\begin{tabular}{|>{\centering\arraybackslash}p{0.3cm}|p{15cm}|}
\hline
ID & Prompt Template\\ \hline
1 & <image> Which object is this person looking at?\textbackslash n A. kettle\textbackslash n B. tulips\textbackslash n C. bunny\textbackslash nIf you don't know, you still must choose one, so make your best guess. There is no need to report intermediate reasoning. Please answer with the option's letter A, B, C directly. \\ \hline
2 & <image> Which object is this person looking at?\textbackslash n A. kettle\textbackslash n B. tulips\textbackslash n C. bunny\textbackslash nThis person is looking at one of the objects.  If you don't know, you still must choose one, so make your best guess. There is no need to report intermediate reasoning. Please answer with the option's letter A, B, C directly. \\ \hline
3 & <image> Which object is this person looking at?\textbackslash n A. kettle\textbackslash n B. tulips\textbackslash n C. bunny\textbackslash nThe objects are placed on the table following the order of tulips, bunny, and kettle, from left to right.  If you don't know, you still must choose one, so make your best guess. There is no need to report intermediate reasoning. Please answer with the option's letter A, B, C directly. \\ \hline
4 & <image> Which object is this person looking at?\textbackslash n A. kettle\textbackslash n B. tulips\textbackslash n C. bunny\textbackslash nThis person is looking at one of the objects. The objects are placed on the table following the order of tulips, bunny, and kettle, from left to right.  If you don't know, you still must choose one, so make your best guess. There is no need to report intermediate reasoning. Please answer with the option's letter A, B, C directly. \\ \hline
5 & <image> Which object is this person looking at?\textbackslash n A. kettle\textbackslash n B. tulips\textbackslash n C. bunny\textbackslash nPlease focus on the eyes. If you don't know, you still must choose one, so make your best guess. There is no need to report intermediate reasoning. Please answer with the option's letter A, B, C directly. \\ \hline
6 & <image> Which object is this person looking at?\textbackslash n A. kettle\textbackslash n B. tulips\textbackslash n C. bunny\textbackslash nThis person is looking at one of the objects. Please focus on the eyes. If you don't know, you still must choose one, so make your best guess. There is no need to report intermediate reasoning. Please answer with the option's letter A, B, C directly. \\ \hline
7 & <image> Which object is this person looking at?\textbackslash n A. kettle\textbackslash n B. tulips\textbackslash n C. bunny\textbackslash nThe objects are placed on the table following the order of tulips, bunny, and kettle, from left to right. Please focus on the eyes. If you don't know, you still must choose one, so make your best guess. There is no need to report intermediate reasoning. Please answer with the option's letter A, B, C directly. \\ \hline
8 & <image> Which object is this person looking at?\textbackslash n A. kettle\textbackslash n B. tulips\textbackslash n C. bunny\textbackslash nThis person is looking at one of the objects. The objects are placed on the table following the order of tulips, bunny, and kettle, from left to right. Please focus on the eyes. If you don't know, you still must choose one, so make your best guess. There is no need to report intermediate reasoning. Please answer with the option's letter A, B, C directly. \\ \hline
9 & <image> Which object is this person looking at?\textbackslash n A. kettle\textbackslash n B. tulips\textbackslash n C. bunny\textbackslash nFirst, locate the person's face. Next, locate their eyes. Finally, trace the line of sight to identify the target object. If you don't know, you still must choose one, so make your best guess. There is no need to report intermediate reasoning. Please answer with the option's letter A, B, C directly. \\ \hline
10 & <image> Which object is this person looking at?\textbackslash n A. kettle\textbackslash n B. tulips\textbackslash n C. bunny\textbackslash nThis person is looking at one of the objects. First, locate the person's face. Next, locate their eyes. Finally, trace the line of sight to identify the target object. If you don't know, you still must choose one, so make your best guess. There is no need to report intermediate reasoning. Please answer with the option's letter A, B, C directly. \\ \hline
11 & <image> Which object is this person looking at?\textbackslash n A. kettle\textbackslash n B. tulips\textbackslash n C. bunny\textbackslash nThe objects are placed on the table following the order of tulips, bunny, and kettle, from left to right. First, locate the person's face. Next, locate their eyes. Finally, trace the line of sight to identify the target object. If you don't know, you still must choose one, so make your best guess. There is no need to report intermediate reasoning. Please answer with the option's letter A, B, C directly. \\ \hline
12 & <image> Which object is this person looking at?\textbackslash n A. kettle\textbackslash n B. tulips\textbackslash n C. bunny\textbackslash nThis person is looking at one of the objects. The objects are placed on the table following the order of tulips, bunny, and kettle, from left to right. First, locate the person's face. Next, locate their eyes. Finally, trace the line of sight to identify the target object. If you don't know, you still must choose one, so make your best guess. There is no need to report intermediate reasoning. Please answer with the option's letter A, B, C directly. \\ \hline
13 & <image> Which object is this person looking at?\textbackslash n A. kettle\textbackslash n B. tulips\textbackslash n C. bunny\textbackslash nFirst, locate the person's face. Next, locate their eyes. Then, enumerate each object in the image to imagine how the eyes would look if the person were looking at that object. Finally, choose the option that best matches the actual eye appearance. If you don't know, you still must choose one, so make your best guess. There is no need to report intermediate reasoning. Please answer with the option's letter A, B, C directly. \\ \hline
14 & <image> Which object is this person looking at?\textbackslash n A. kettle\textbackslash n B. tulips\textbackslash n C. bunny\textbackslash nThis person is looking at one of the objects. First, locate the person's face. Next, locate their eyes. Then, enumerate each object in the image to imagine how the eyes would look if the person were looking at that object. Finally, choose the option that best matches the actual eye appearance. If you don't know, you still must choose one, so make your best guess. There is no need to report intermediate reasoning. Please answer with the option's letter A, B, C directly. \\ \hline
15 & <image> Which object is this person looking at?\textbackslash n A. kettle\textbackslash n B. tulips\textbackslash n C. bunny\textbackslash nThe objects are placed on the table following the order of tulips, bunny, and kettle, from left to right. First, locate the person's face. Next, locate their eyes. Then, enumerate each object in the image to imagine how the eyes would look if the person were looking at that object. Finally, choose the option that best matches the actual eye appearance. If you don't know, you still must choose one, so make your best guess. There is no need to report intermediate reasoning. Please answer with the option's letter A, B, C directly. \\ \hline
16 & <image> Which object is this person looking at?\textbackslash n A. kettle\textbackslash n B. tulips\textbackslash n C. bunny\textbackslash nThis person is looking at one of the objects. The objects are placed on the table following the order of tulips, bunny, and kettle, from left to right. First, locate the person's face. Next, locate their eyes. Then, enumerate each object in the image to imagine how the eyes would look if the person were looking at that object. Finally, choose the option that best matches the actual eye appearance. If you don't know, you still must choose one, so make your best guess. There is no need to report intermediate reasoning. Please answer with the option's letter A, B, C directly. \\ \hline
\end{tabular}
\label{table: prompts}
\end{table}
\end{strip}

\subsection{Human Response Collection Details}
\label{subsection: human_survey}
We used Prolific to recruit 59 participants around the globe who are fluent in English and use Desktop browsers to access our survey (created using JsPsych; developed by \citealt{jspsych}).
The pilot stimuli are split into 20 predetermined stimulus lists with 45 test stimuli per list.
There are additionally 7 attention-check stimuli with \texttt{\#Objects=2} and \texttt{Proximity=1} while covering a range of \texttt{Objects}, all three \texttt{View}s, and both \texttt{Gazer}s.
Each participant was assigned to one of the stimulus lists and received a mix of 45 test stimuli and 7 attention checks with random presentation order.
Although some participants failed at least one attention check, we did not exclude them from analysis.

\paragraph{Attention Checks}
The attention checks are the same across participants and are meant to be the simplest cases that a focused person should not fail.
Indeed, within participants who are correct on all attention checks, the non-attention-check questions with \texttt{\#Objects=2} and \texttt{Proximity=1} (meaning that they are as difficult as the attention checks) have a mean accuracy of around 99.30\%.

\begin{figure}
\centering
\includegraphics[alt={}, width=0.8\linewidth]{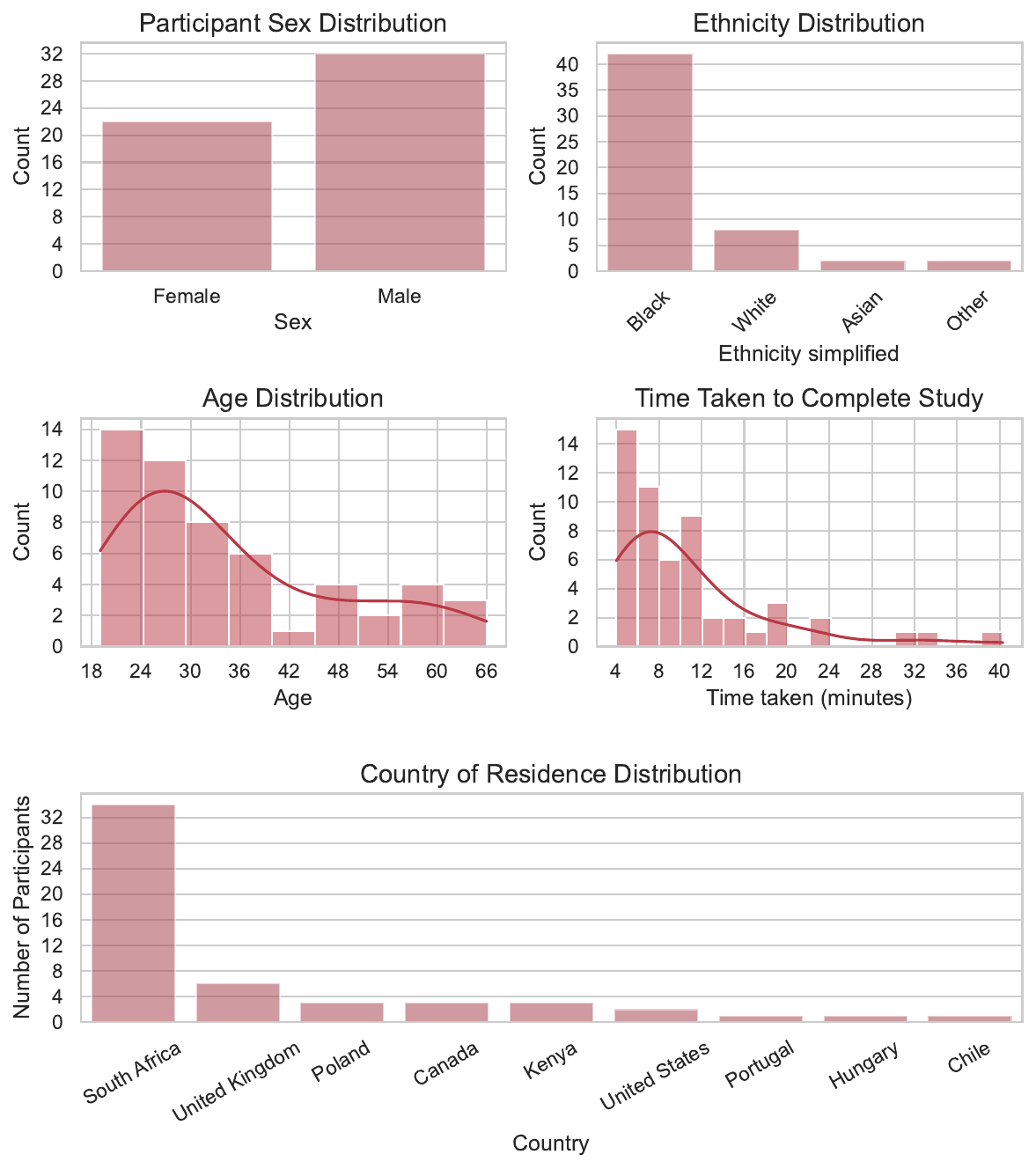}
\caption{The demographics. }
\label{fig: demographics}
\end{figure}

\paragraph{Human Participant Demographics}
See Fig.~\ref{fig: demographics} for the demographics of the 59 participants.
Demographic data of some participants is missing and hence not plotted.
All participants are paid an hourly rate of 12 USD regardless of their residence.

\begin{figure}
\centering
\includegraphics[alt={}, width=\linewidth]{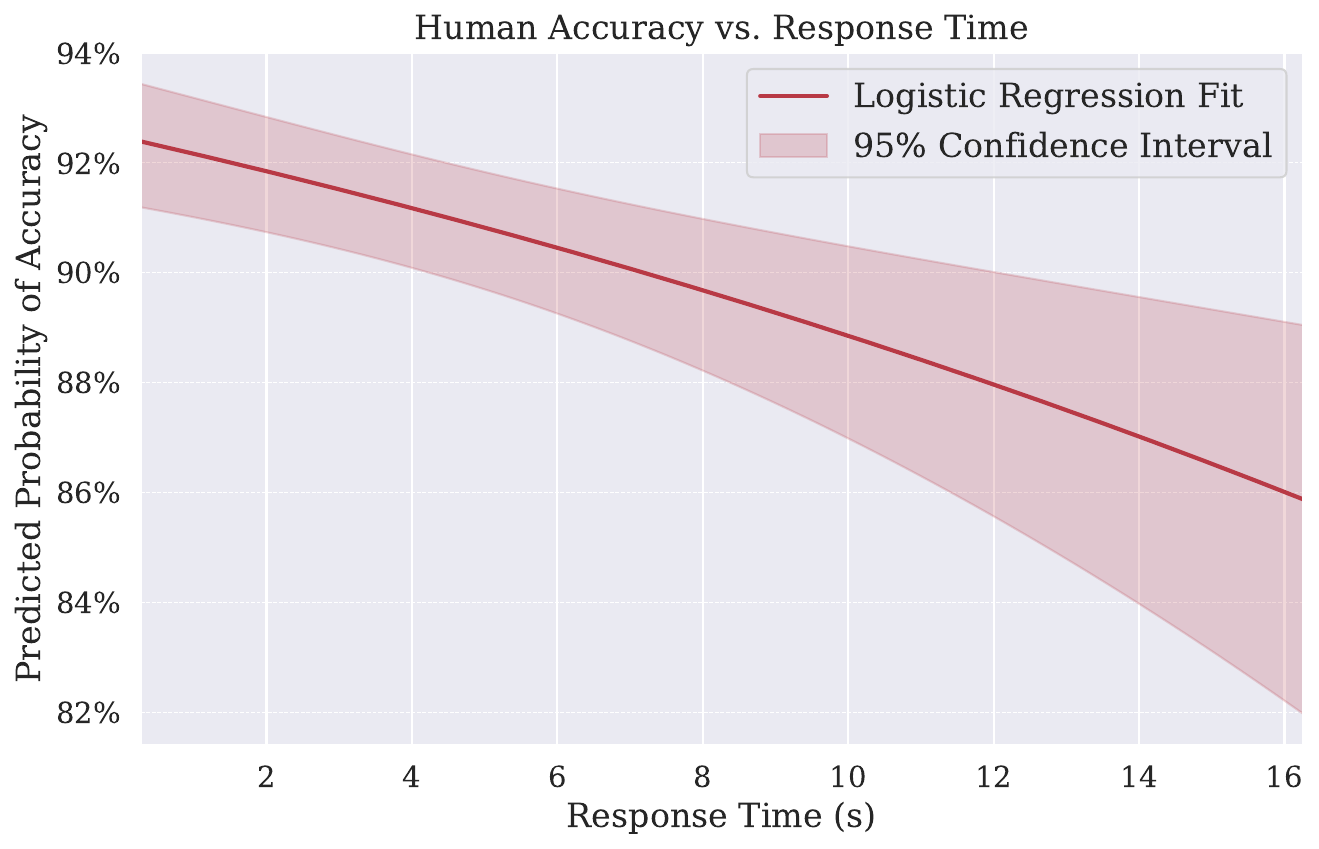}
\caption{The prediction made by a logistic regression model (response time shown here as the 99th percentile range). }
\label{fig: human_accuracy_vs_rt}
\end{figure}

\paragraph{Human Accuracy vs. Response Time}
See Fig.~\ref{fig: human_accuracy_vs_rt}.
We submitted the response time details for every trial of valid participants (who passed all attention checks).
The higher the response time of the trial is, the lower the likelihood of getting the correct response ($p < .001$).
This indicates that people tend to spend more time on harder cases and still tend to fail on them.

\begin{figure}[H]
\begin{subfigure}{\linewidth}
\centering
\includegraphics[alt={}, width=0.85\linewidth]{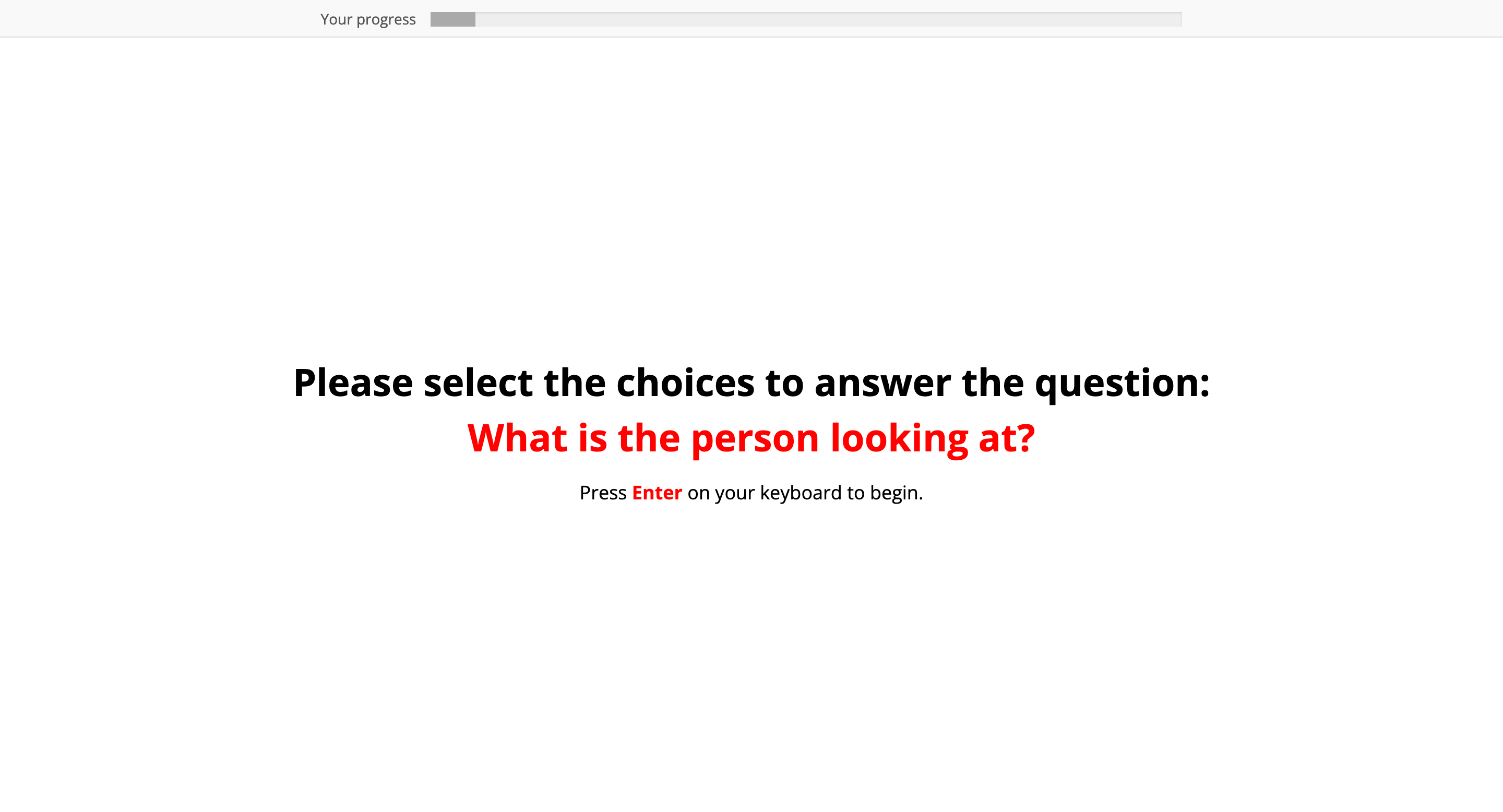}
\caption{The first instruction page after being put into full-screen mode.}
\end{subfigure}
\\
\\
\begin{subfigure}{\linewidth}
\centering
\includegraphics[alt={}, width=0.85\linewidth]{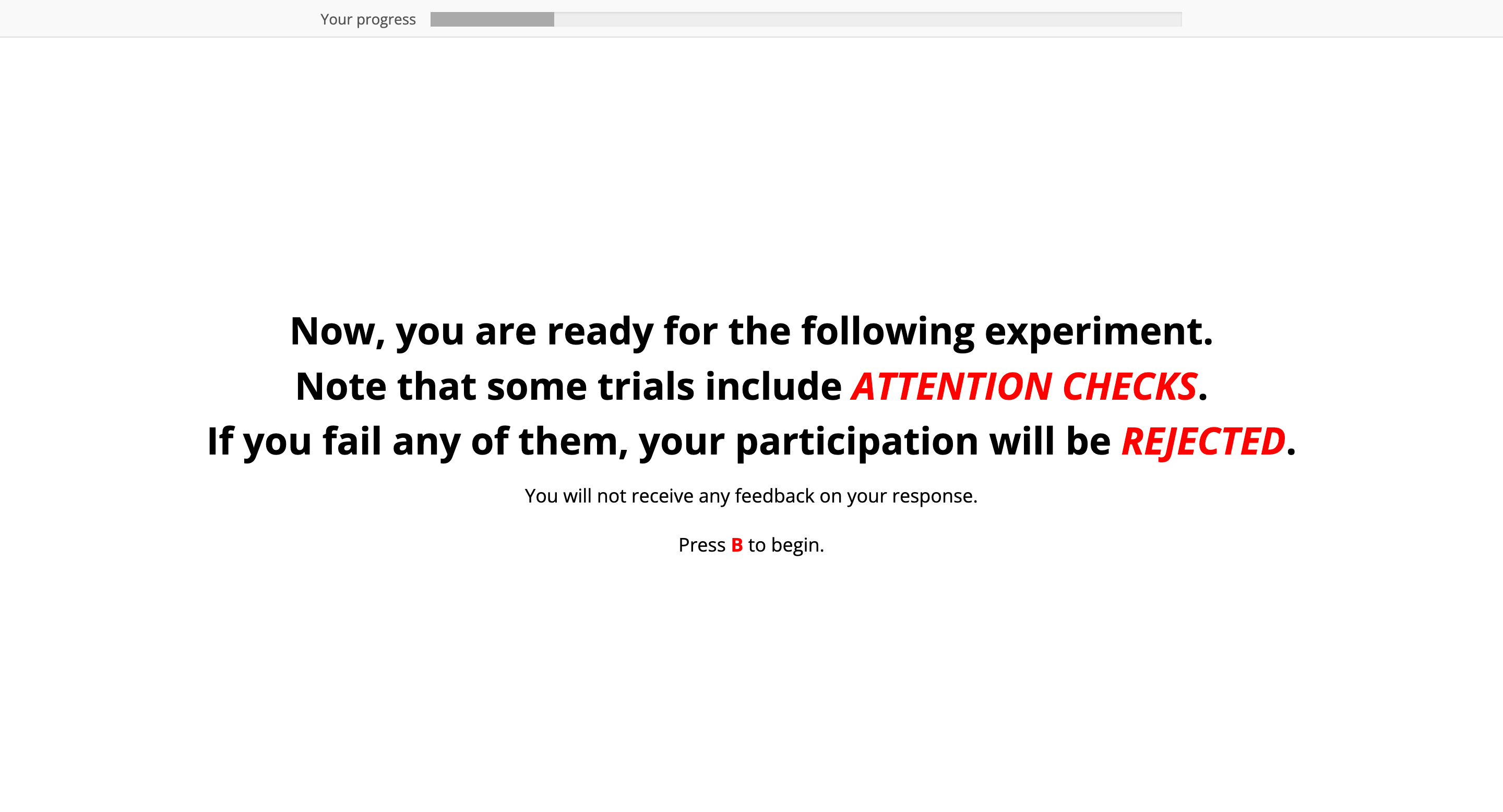}
\caption{The second instruction page explains the existence of attention checks.}
\end{subfigure}
\\
\\
\begin{subfigure}{\linewidth}
\centering
\includegraphics[alt={}, width=0.85\linewidth]{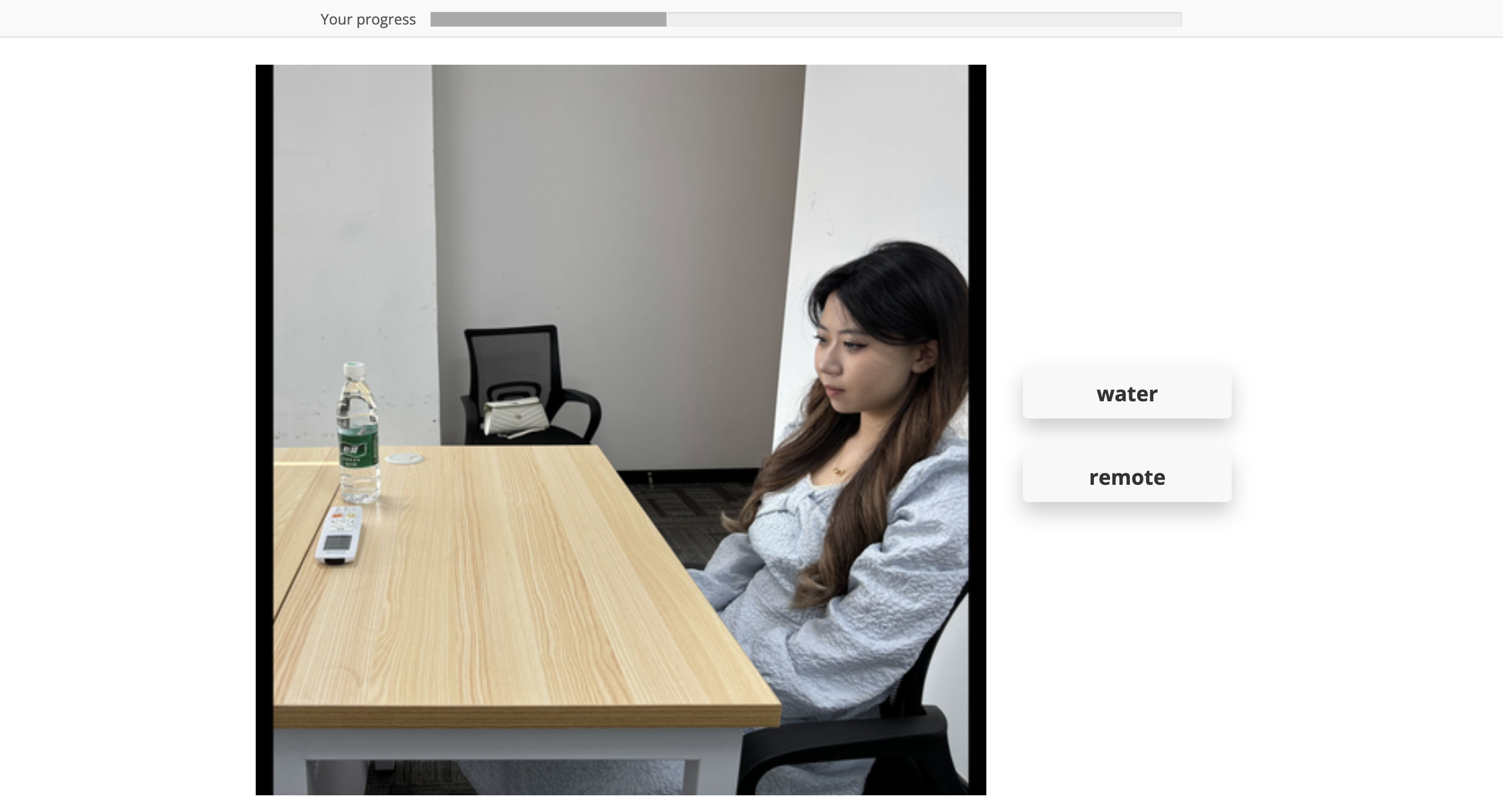}
\caption{An example of the question pages where participants click one of the buttons to make their choice and proceed to the next question.}
\end{subfigure}

\caption{
Screenshots from the human survey. }
\label{fig: human_survey}
\end{figure}

\paragraph{Details of the Survey}
See Fig.~\ref{fig: human_survey} for screenshots of the survey.
The participants click a button to enter full-screen mode and start reading instructions (they have to press different keys on the keyboard to ensure they read the instructions).
Then, they go through three practice trials based on 3 of the seven attention checks with correctness feedback (in the form of a check mark or a cross mark with no sound).
After that, they complete the 45 (test stimuli) + 7 (attention check) questions with a progress bar to motivate them, as the questions are relatively easy, but no correctness feedback is available.
The stimuli in the three practice trials will appear again as attentional checks that look the same as the other trials.
Four other pre-determined attention checks will also appear.
Participants do not know which are attention checks, and they are warned to try their best for all questions.
Even if they failed attention checks, they were still paid, regardless of the warning.
Participants cannot go back and forth to make changes, but they can take as long as they want to answer the questions.

\subsection{Pre-pilot Results}
\label{subsection: ranking}
\begin{figure}[H]
\centering
\includegraphics[alt={}, width=0.88\linewidth]{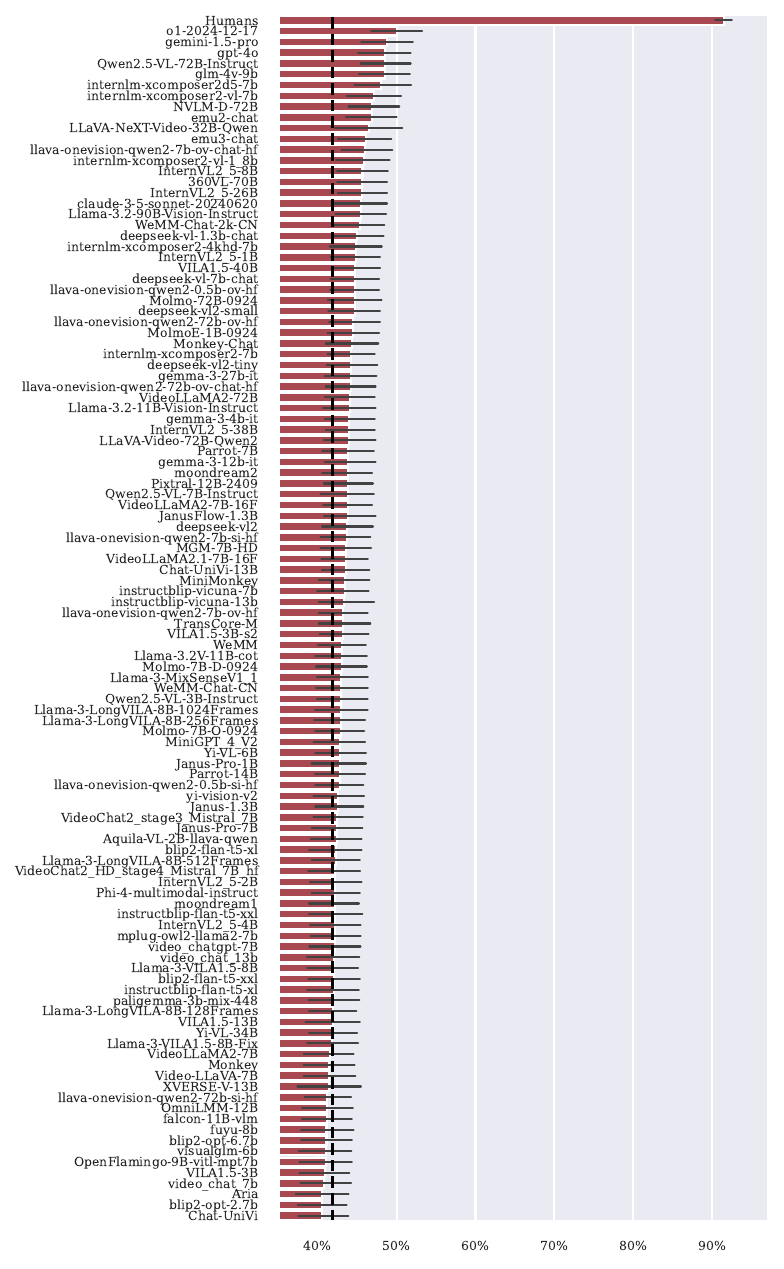}
\caption{Full comparison of the overall accuracy of Humans and VLMs. 95\% CIs are drawn horizontally, while the random guessing baseline of 42\% is drawn vertically.}
\label{fig: ranking}
\end{figure}

See Fig.~\ref{fig: ranking}. 
Note that responses collected in the pilot for the 5 selected VLMs are excluded to preserve equal sample sizes among VLMs for fair comparison.

\subsection{Scaling Does Not Help}
\label{subsection: scaling}
\begin{figure}
  \centering
  \includegraphics[width=\linewidth]{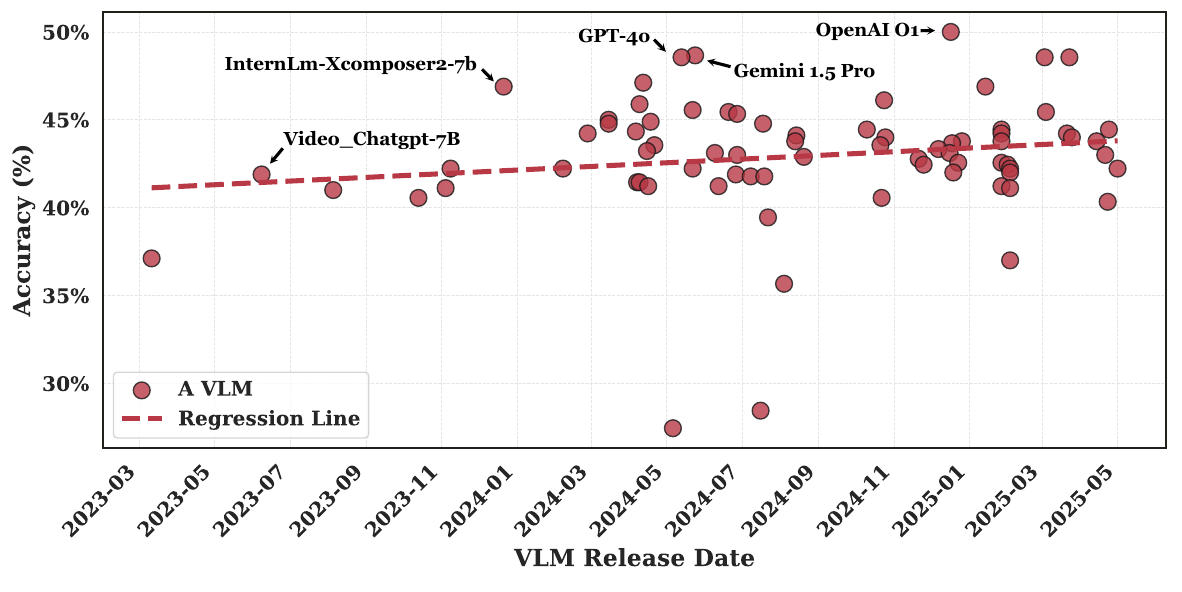}
  \caption{No strong linear relation between VLM accuracy and release date was found.}
  \label{fig: ranking_time}
\end{figure}

\begin{figure}
\centering

\includegraphics[width=\linewidth]{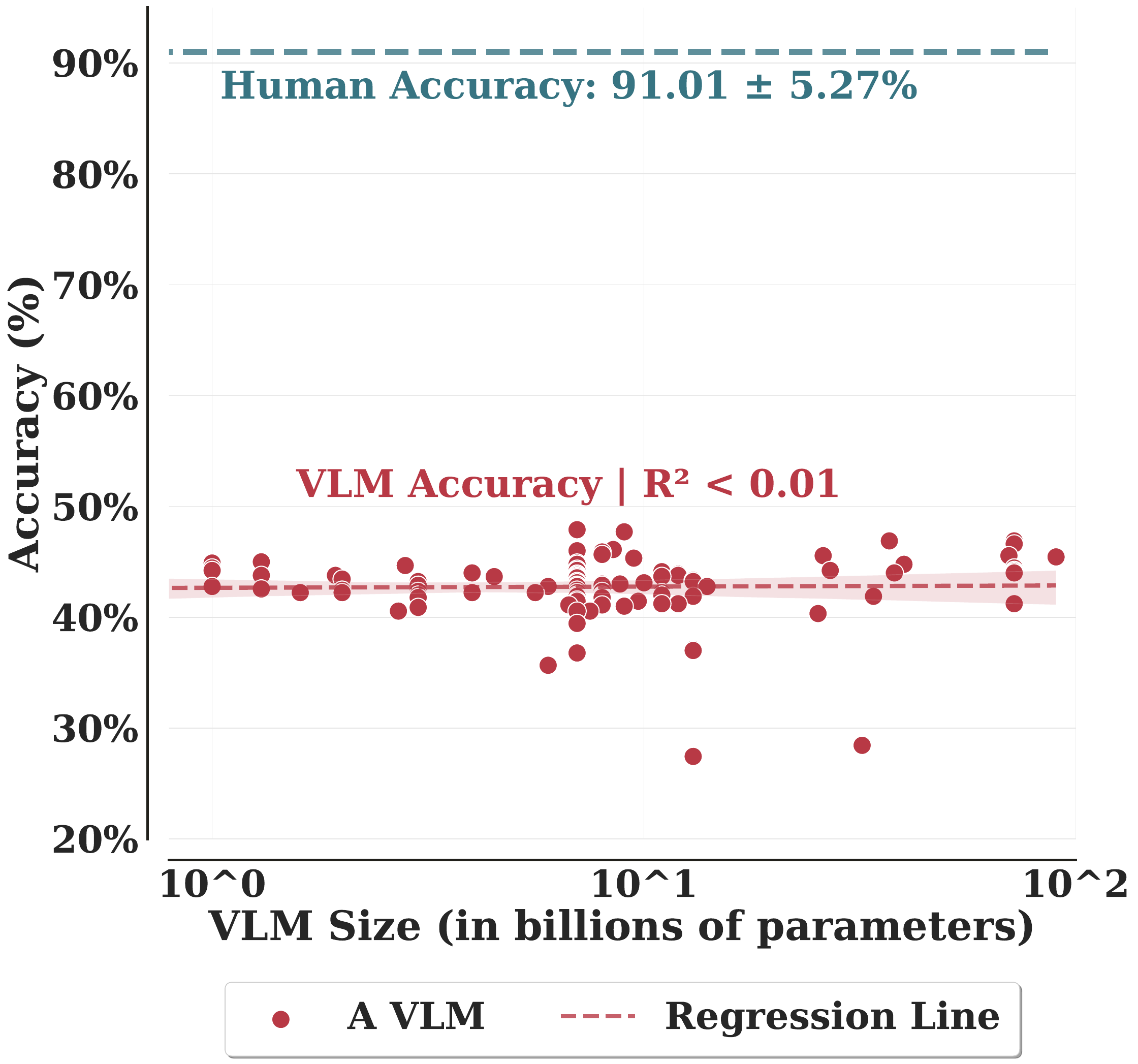}
\caption{The higher the accuracy, the better. Accuracy does not improve as the number of parameters increases. The 95\% confidence intervals for linear regression are drawn as shaded areas. Standard deviations are reported for variables drawn as horizontal lines.}
\label{fig: ranking_size}

\end{figure}

We collected VLM release date information for 75 of the 111 VLMs (excluding smaller-size versions when the full-size versions are present) and size estimates for 106 VLMs (excluding outliers with estimated size larger than 100 billion parameters).
They cannot linearly predict accuracy ($R^2 < 0.03, 0.01$ respectively), as shown in Fig.~\ref{fig: ranking_time} and Fig.~\ref{fig: ranking_size}, respectively.

\subsection{Main Study Mixed-Effects Modeling Details}
\label{subsection: appendix_modeling}
We fitted mixed-effects logistic models in R (Version 4.4.3) using the lme4 package (Version 1.1-38; \citealt{lme4}).

\subsubsection{Proximity and View Effects}
\begin{figure}
\begin{subfigure}{\linewidth}
\includegraphics[alt={}, width=\linewidth]{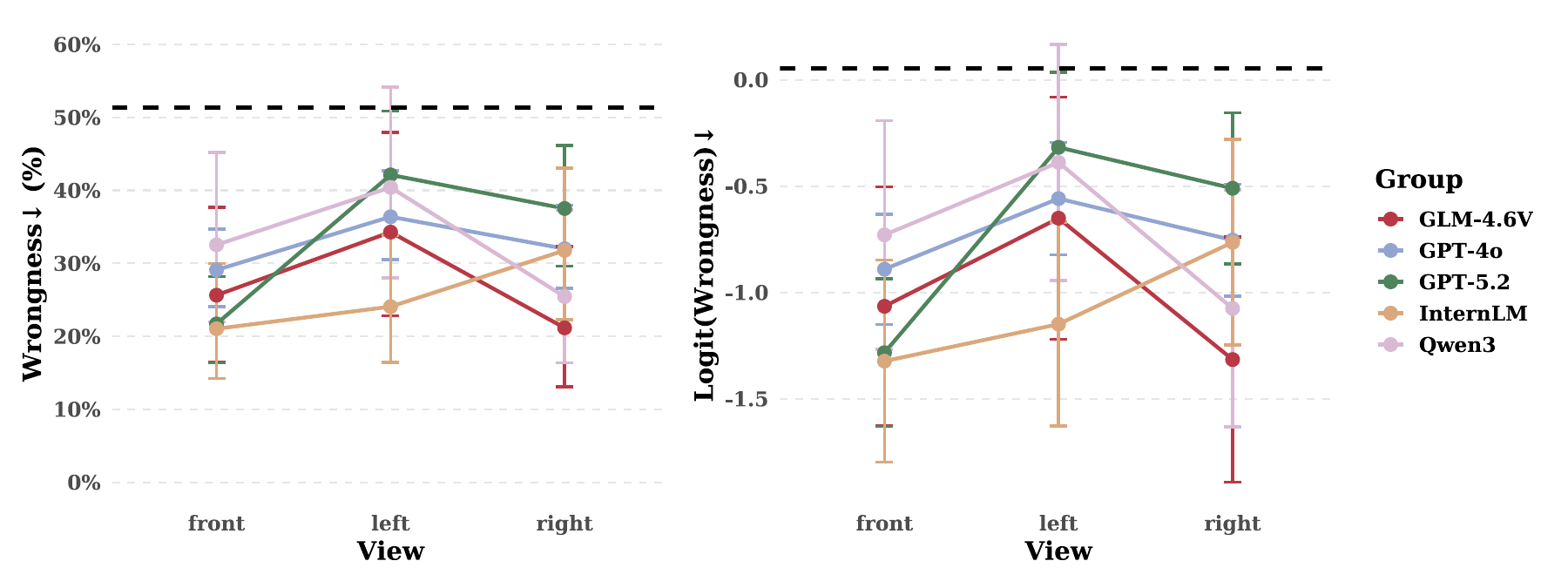}
\caption{\textit{View effect}}
\label{fig: angle_effect}
\end{subfigure}
\\
\begin{subfigure}{\linewidth}
\includegraphics[alt={}, width=\linewidth]{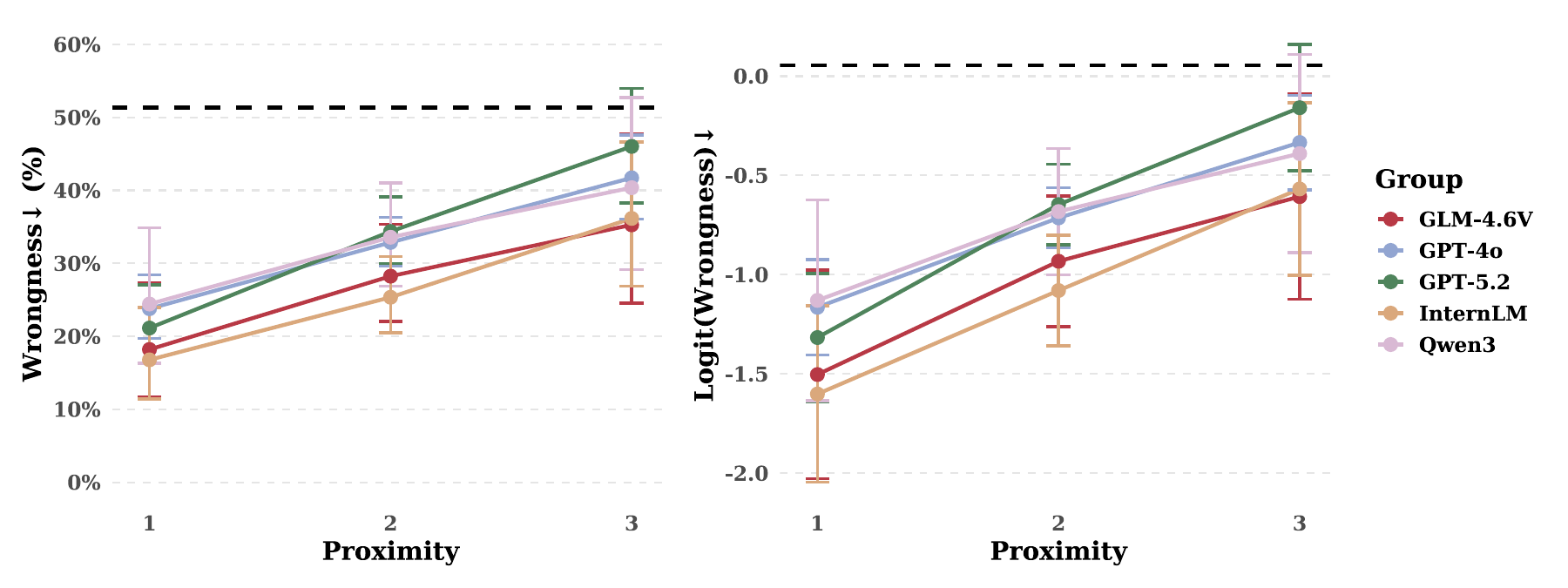}
\caption{\textit{Proximity effect}}
\label{fig: proximity_effect}
\end{subfigure}
\\
\begin{subfigure}{\linewidth}
\includegraphics[alt={}, width=\linewidth]{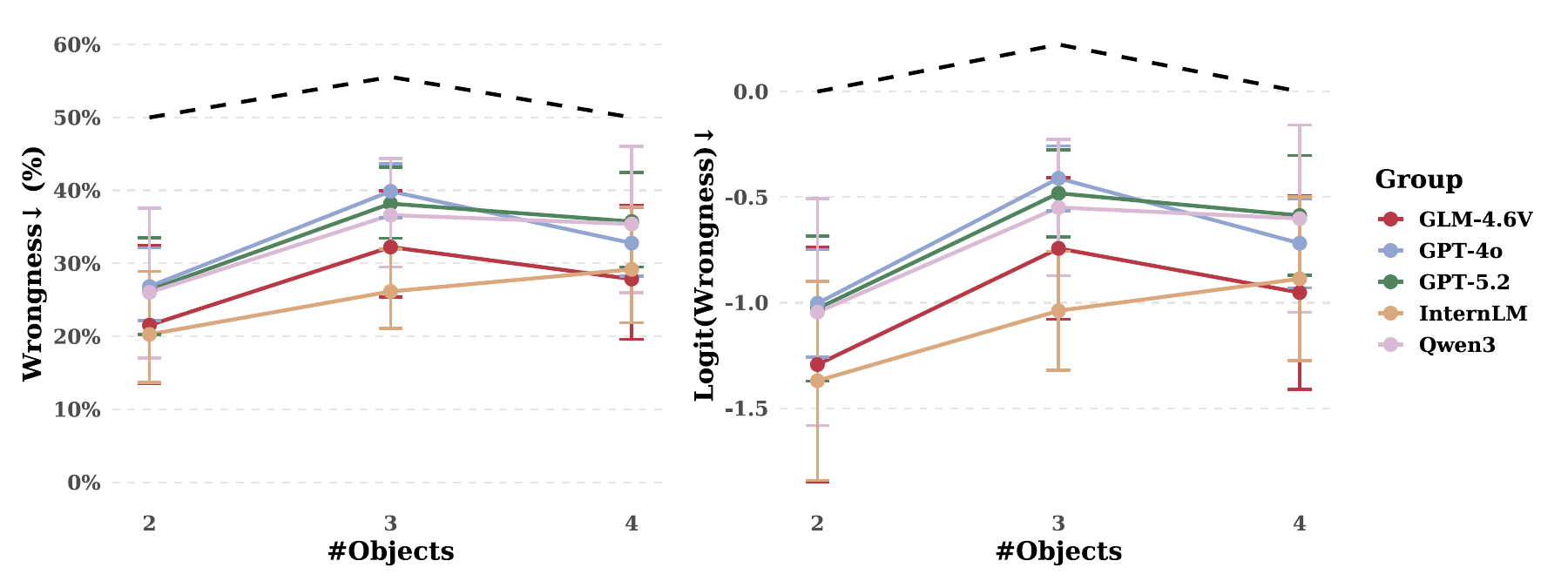}
\caption{\textit{Number of Candidate effect, exploratory}}
\label{fig: n_candidates_effect}
\end{subfigure}

\end{figure}

Figure~\ref{fig: angle_effect} shows the estimated marginal means obtained by fitting a mixed-effects model for each group.
Dashed lines are the random-guessing baselines. 
The figures on the left depict variable relations with the wrongness in the zero-to-one space, while the figures on the right depict them in the logit space.
Note that averaging is always performed in the logit space, as this is part of the reason why the link transformation is used in the first place.
No significant difference between marginal means for \texttt{View=left} and \texttt{right}.

\subsubsection{Choosing \texttt{HeadTarget} More Often than \texttt{GazeTarget} (Test 2.1)}
\paragraph{Model}
Take results in the Incongruent Condition with \texttt{\#Objects}>2 and \texttt{Choice} being either \texttt{HeadTarget} or \texttt{GazeTarget}.
Fit a binary logistic GLMM $\mathrm{logit(IsHeadTarget)} \sim \beta_0$ with aforementioned random effects for each VLM, where \texttt{IsHeadTarget} is the binary variable of whether or not \texttt{Choice} equals \texttt{HeadTarget}.
\paragraph{Results}
The intercept $\beta_0$ represents the ratio of choosing the \texttt{HeadTarget} (when \texttt{Choice} being either \texttt{HeadTarget} or \texttt{GazeTarget}) for a stimulus that induces "median" level head bias.
For GPT-5.2, the intercept $\beta_0$ is positive ($p<1e-4$), indicating a positive answer to the test question.
Indeed, it chose \texttt{HeadTarget} 62\% of the time when \texttt{Proximity}=1.
This rate returns to 52\% when \texttt{Proximity}=3, which is not surprising because when these two choices are spatially closer, the head bias can be less salient.
The test is taken to be positive for GPT-5.2.
For Qwen3, it chose \texttt{HeadTarget} 57\% of the time when \texttt{Choice} was either \texttt{HeadTarget} or \texttt{GazeTarget}.
As the bimodal pattern made the statistical model not reliable for Qwen3, its test result is taken to be positive, per a t-test ($p<1e-3$).
The model is similarly unreliable for GLM, but its t-test result is negative ($51.6\%, p=0.15$).
This test is negative for GPT-4o and InternLM.

\subsubsection{Respond the Same When \texttt{HeadTarget} Stays the Same (Test 2.2)}
\paragraph{Model}
Take results in the Congruent and Incongruent Conditions.
Fix a combination of (\texttt{Objects}, \texttt{Proximity}, \texttt{View}, \texttt{HeadTarget}, \texttt{PromptID}), vary only \texttt{GazeTarget} (2-4 levels) and see whether \texttt{Choice} stays the same.
Calculate the number of combinations where this is the case.
\paragraph{Results}
GPT-5.2 chose the same option across those 2 to 4 stimuli that only differ in the eye appearance around 58\% of the time, much higher than chance ($chance\approx20\%, p<3e-16$, combination-level directional one-sample t-test).
Specifically, around 29\% of the time, it always chose \texttt{HeadTarget} across stimuli within a combination, a rate again much higher than chance ($chance\approx10\%, p<3e-16$), leading to a positive result.
Similar results for all other VLMs.

\subsubsection{Perform Worse When \texttt{HeadTarget} $\ne$ \texttt{GazeTarget} (Test 2.3)}
\paragraph{Model}
Fit $\mathrm{logit(IsCorrect)} \sim \beta_0 + \mathrm{logit}(\mathbb{E}(\mathrm{IsCorrect})) +\mathrm{View} \times \mathrm{Proximity} \times  \mathrm{{\#Objects}} \times \mathrm{Condition}$ with aforementioned random effects, where \texttt{IsCorrect} is the binary variable of whether or not \texttt{Choice} equals \texttt{GazeTarget}.
\paragraph{Results}
GPT-5.2 solved 45\% of the problems in the Congruent Condition, but only 28\% in the Incongruent Condition, a strong positive result.
Similar results for all other VLMs.

\subsubsection{Distance(\texttt{Choice}, \texttt{GazeTarget}) Increases as Distance(\texttt{HeadTarget}, \texttt{GazeTarget}) Increases (Test 2.4)}
\paragraph{Model}
Fit $\mathrm{logit} (\mathrm{Wrongness}) \sim  \beta_0 + \mathrm{logit}(\mathbb{E}(\mathrm{Wrongness})) + \mathrm{View} \times \mathrm{Proximity} \times  \mathrm{{\#Objects}} \times \mathrm{DistEyeHead}$ with aforementioned random effects.
\paragraph{Results}
For GPT-5.2, the marginal slope indicates that each time DistEyeHead increases by 1, the odds of \texttt{Wrongness} increase by 10 percent on average ($p<1e-13$).
Similar results for all other VLMs.
This effect is stronger for stimuli with smaller \texttt{Proximity}, consistently for all these VLMs.

\subsubsection{Distance(\texttt{HeadTarget}, \texttt{Choice}) Positively Correlates with Distance(\texttt{HeadTarget}, \texttt{GazeTarget}) (Test 3)}
\paragraph{Model}
Fit $\mathrm{logit} (\mathrm{DistToHeadTarget}) \sim  \beta_0 + \mathrm{logit}(\mathbb{E}(\mathrm{DistToHeadTarget})) + \mathrm{View} \times \mathrm{Proximity} \times  \mathrm{{\#Objects}} \times \mathrm{DistEyeHead}$ with aforementioned random effects, where \texttt{DistToHeadTarget} is the relative distance between \texttt{Choice} and \texttt{HeadTarget}, defined in a way similar to \texttt{Wrongness}.
\paragraph{Results}
There is no such effect for Qwen3.
For GPT-5.2, the marginal slope indicates that when \texttt{Proximity}=1, as DistEyeHead increases by 1, the odds of \texttt{DistToHeadTarget} increase by 5 percent on average ($p=0.028$).
This effect becomes marginal when averaging across different values of \texttt{Proximity} ($p=0.054$), so we still take GPT-5.2 to have a negative answer to this question.
For GPT-4o, InternLM, and GLM, this test is positive.

\subsubsection{Whether Frequency of Choosing \texttt{HeadTarget} Changes Within the Incongruent Condition (Test 4)}
\paragraph{Model}
Take results in the Congruent and Incongruent Conditions with \texttt{\#Objects} greater than 2. 
Fit $\mathrm{logit} (\mathrm{IsHeadTarget}) \sim  \beta_0 + \mathrm{logit}(\mathbb{E}(\mathrm{IsHeadTarget})) + \mathrm{View} \times \mathrm{Proximity} \times  \mathrm{{\#Objects}} \times \mathrm{DistEyeHead}$ with aforementioned random effects.
Crucially, \texttt{DistEyeHead} is treated as categorical. 
Separately looking at \texttt{\#Object}=3 and 4.
\paragraph{Inference}
If the marginal means of \texttt{IsHeadTarget} for \texttt{DistEyeHead}=1 and 2 when \texttt{\#Object}=3 (or 1, 2, and 3 when \texttt{\#Object}=4) are not significantly different from each other, but are all significantly lower than when \texttt{DistEyeHead}=0 (the Congruent Condition), then the type 5 is supported. 
Otherwise, if the marginal mean decreases overall, then the type 4 is supported.
\paragraph{Results}
GPT-5.2 marginal means of \texttt{IsHeadTarget} did not change as \texttt{DistEyeHead} changes.
Because of this lack of support for either type 4 or 5 and its negative Test 3 result, we categorized it as type 3.
It is similarly the case for Qwen3 and GLM.
Curiously, GLM also showed a positive result of Test 3, meaning that it is responsive to eye appearance, but not in a way described by types 4 or 5.
In contrast, for GPT-4o, the marginal means of \texttt{IsHeadTarget} dropped as \texttt{DistEyeHead} goes from 1 to 2 when \texttt{\#Object}=3, as well as when \texttt{DistEyeHead} goes from 2 to 3 when \texttt{\#Object}=4 ($p=0.007, 0.034$, respectively).
This makes GPT-4o a prototypical member of type 4.
For InternLM, it only dropped as \texttt{DistEyeHead} goes from 1 to 2 when \texttt{\#Object}=3 ($p<1e-3$) but not for \texttt{\#Object}=4.
Since no difference between \texttt{DistEyeHead}=0 and 1 was found for both values of \texttt{\#Object}, InternLM is more likely a member of type 4 than 5.

\subsubsection{Other Exploratory Analysis}
\paragraph{Ability to Ignore Background Distractors}
VLMs can ignore background distractors.
Main study stimuli were collected in two locations–a blue room and a white room (see Fig.~\ref{fig: decisiontree} for examples from each).
There are no background distractors for blue room stimuli and only a decorative wall canvas for white room stimul.
Since no performance difference is detected between these two cases, we effectively presented scenes with minimal visual distractions, and VLMs can ignore such a simple distractor.
Additionally, in the pilot stimuli, there are only cluttered background distractors when \texttt{View=right}.
Since there is almost no View effect found in the pilot (except for Gemini 1.5 Pro), VLMs are not greatly distracted by a more cluttered background either.

\subsection{Finetuning Dynamics Across 5 Runs}

\paragraph{Variants of GazeLLE}
There are 4 variants of GazeLLE.
Its backbone can be DINOv2 ViT-L or DINOv2 ViT-B.
We can supply the bounding boxes for faces or not (annotated by Moondream3-preview and all manually checked).
We evaluated all and showcased the one that performed best—GazeLLE DINOv2 ViT-L with annotated face bounding boxes, while other variants performed pretty well too.

\paragraph{GazeLLE Finetuning Results}
We visualize the training trajectories for each of the five independent runs across all conditions
Most importantly, the improvement in the Incongruent condition is reliable across different runs.
\begin{figure}[h!]
\centering

\includegraphics[width=\columnwidth]{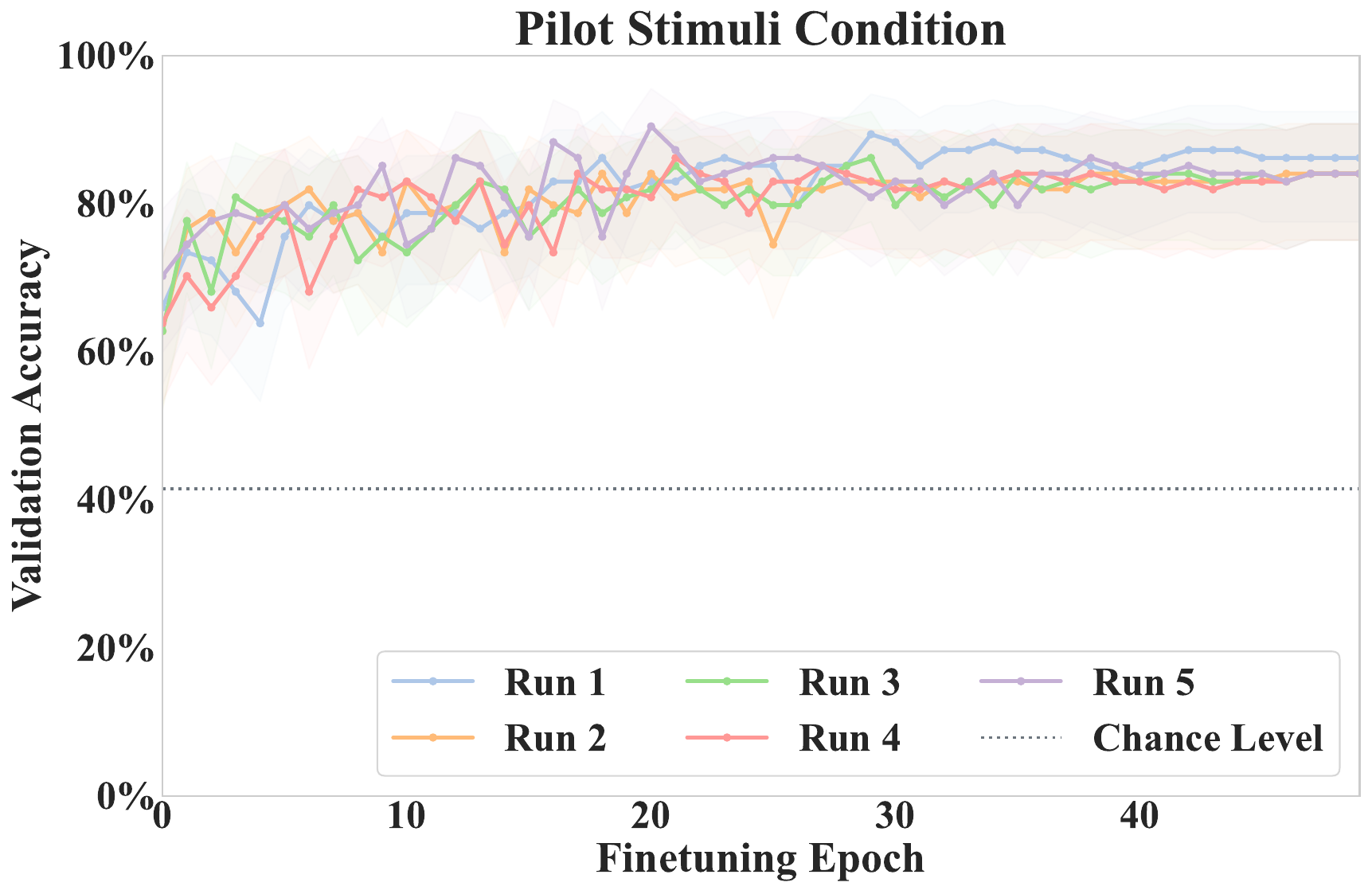}

\vspace{0.5em}

\includegraphics[width=\columnwidth]{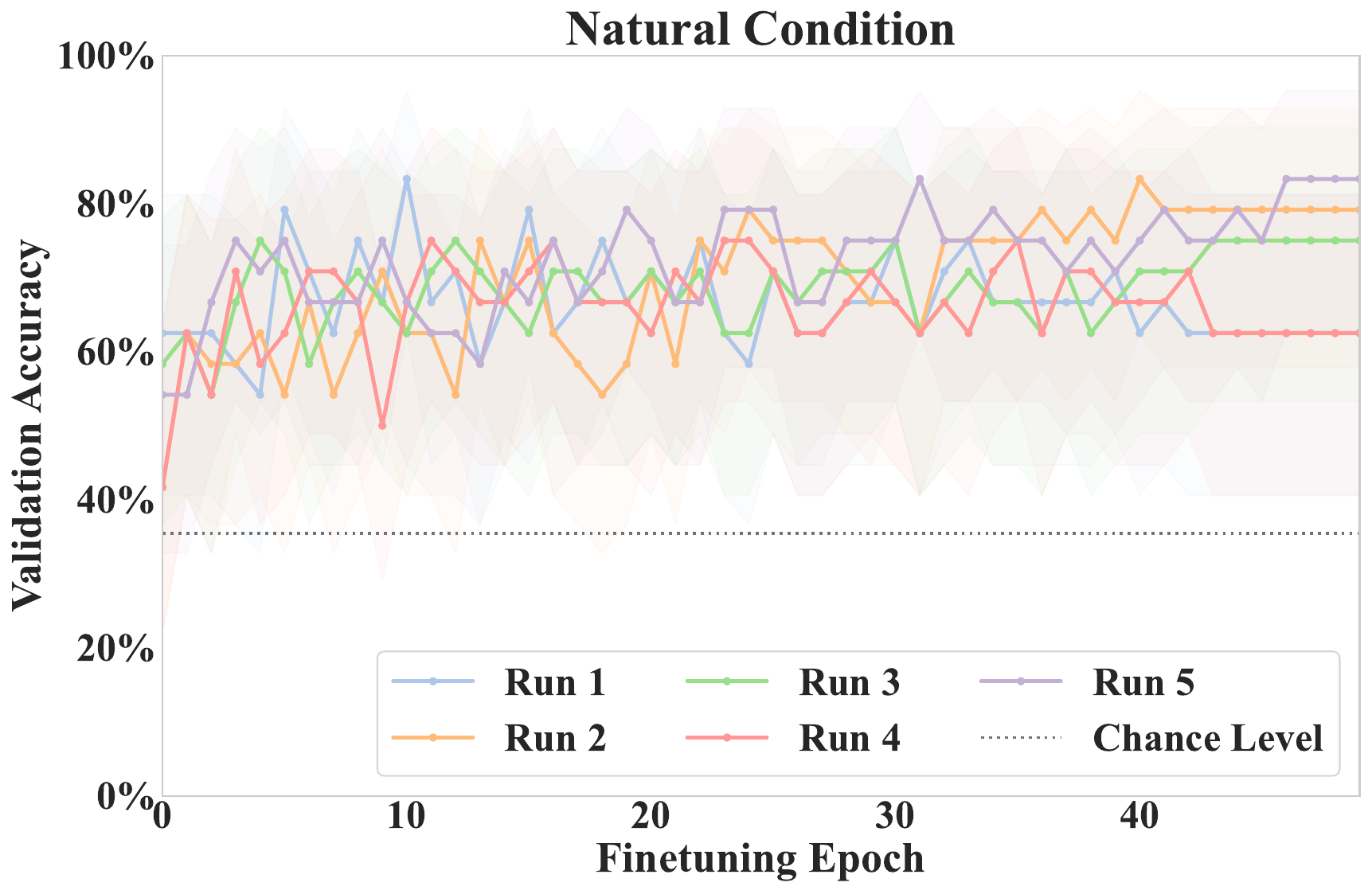}

\vspace{0.5em}

\includegraphics[width=\columnwidth]{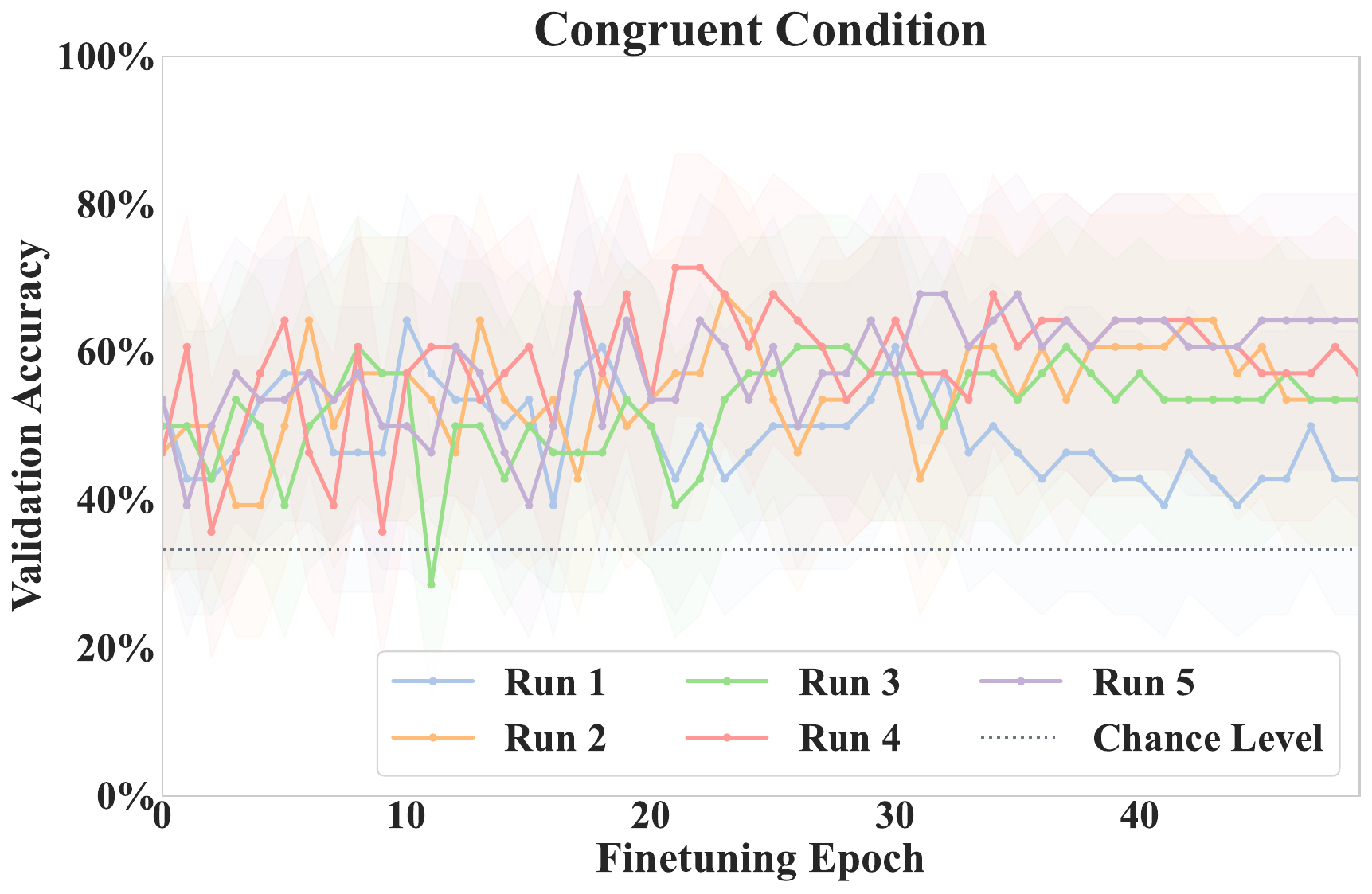}

\vspace{0.5em}

\includegraphics[width=\columnwidth]{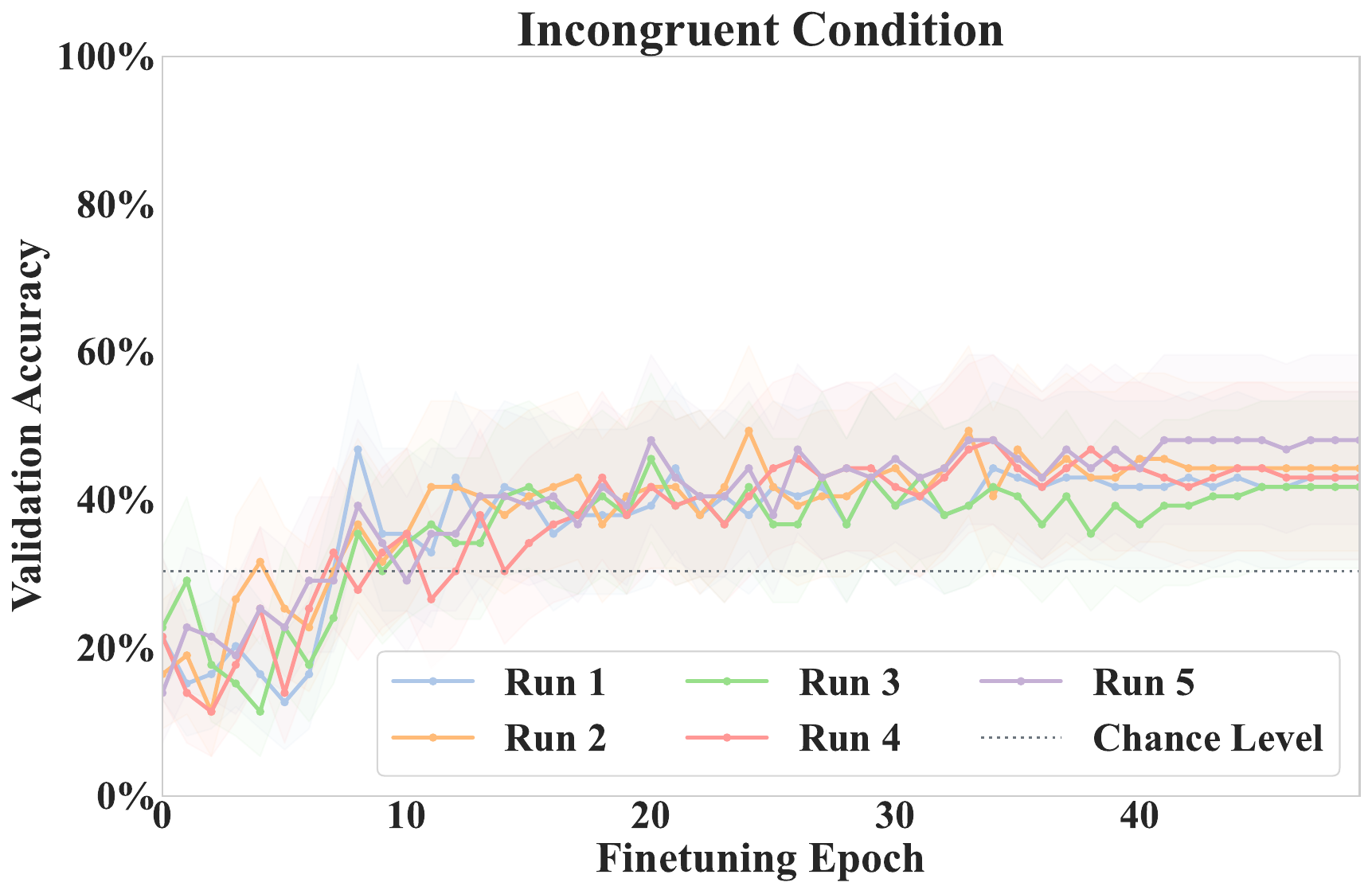}
\caption{Validation accuracy across fine-tuning epochs for five independent runs across conditions.}
\label{fig: gazelle_dynamics_per_run}
\end{figure}

\end{document}